\documentclass[10pt,letterpaper]{article}

\usepackage{framed}
\usepackage{cogsci_template}
\usepackage{cogsci}
\usepackage{float} 
\usepackage{ifsym}

\colorlet{shadecolor}{gray!25}

\usepackage{tcolorbox} 
\tcbuselibrary{breakable} 

\newenvironment{prompt}{
  \par\medskip 
  \noindent
  \begin{tcolorbox}[
    breakable, 
    left=5mm, 
    right=5mm, 
    boxsep=2mm, 
    colback=gray!20, 
    colframe=gray!50, 
    width=\columnwidth, 
    enlarge left by=0mm, 
    enlarge right by=0mm 
  ]
  \fontfamily{pcr}\selectfont 
}{
  \end{tcolorbox}
  \par\medskip 
}

\usepackage{tikz}
\usepackage{bbm}
\usepackage{diagbox}
\usepackage{pifont}
\usepackage[hang,flushmargin]{footmisc}
\usepackage{textcomp,scalerel}
\hypersetup{
    citecolor=gray
}
\SetKwRepeat{Do}{do}{while}%
\acrodef{llm}[LLM]{Large Language Model}
\acrodef{tom}[ToM]{Theory of Mind}
\acrodef{ir}[IR]{\textit{Inverse Reasoning}}
\acrodef{iip}[IIP]{\textit{Inverse Inverse Planning}}
\acrodef{agi}[AGI]{Artificial General Intelligence}
\acrodef{asi}[ASI]{Artificial Social Intelligence}

\def\nFar{\textit{Avoidant}}
\def\nBest{\textit{Hybrid}}
\def\nMisld{\textit{Reversed}}
\def\nShort{\textit{Shortest}}

\def\nDirect{Intermediate}
\def\nFinal{Last}
\def\nTurn{Previsited}
\def\Prob{\mathbb{P}}

\def\aa{i}

\newcommand*\circled[1]{\tikz[baseline=(char.base)]{%
    \node[shape=circle,draw,inner sep=2pt] (char) {#1};}%
}

\cogscifinalcopy

\title{Evaluating and Modeling Social Intelligence:\\A Comparative Study of Human and AI Capabilities}

\author{
    \begin{tabular}{c c c c c}
        \small\bf Junqi Wang$^{\star,1}$ & \small\bf Chunhui Zhang$^{\star,1}$ & \small\bf Jiapeng Li$^{1,2}$ & \small\bf Yuxi Ma$^{1}$  & \small\bf Lixing Niu$^{1, 3}$\\
        \normalfont wangjunqi@bigai.ai & \normalfont zhangchunhui@bigai.ai & \normalfont lijiapeng@stu.xjtu.edu.cn & \normalfont mayuxi@bigai.ai & \normalfont lxniu@stu.pku.edu.cn
    \end{tabular}
    \\
    \begin{tabular}{c c c c}
        \small\bf
 Jiaheng Han$^{1,3}$ & \small\bf Yujia Peng$^{1,4,5,\,\textrm{\Letter}}$ & \small\bf Yixin Zhu$^{4,\,\textrm{\Letter}}$ & \small\bf Lifeng Fan$^{1,\,\textrm{\Letter}}$\\
        \normalfont hanjiaheng@pku.edu.cn & \normalfont yujia\_peng@pku.edu.cn & \normalfont yixin.zhu@pku.edu.cn & \normalfont lifengfan@bigai.ai
    \end{tabular}\vspace{3pt}
    \\\footnotesize $\star$ equal contributors\quad{}$\textrm{\Letter}$ corresponding authors\quad{}
    \footnotesize $^1$ State Key Laboratory of General Artificial Intelligence, BIGAI\\
    \footnotesize $^2$ National Key Laboratory of Human-Machine Hybrid Augmented Intelligence, Xi'an Jiaotong University\\
    \footnotesize $^3$ School of Intelligence Science and Technology, Peking University\quad{}
    \footnotesize $^4$ Institute for Artificial Intelligence, Peking University\\
    \footnotesize $^5$ School of Psychological and Cognitive Sciences and Beijing Key Laboratory of Behavior and Mental Health, Peking University\\
    \vspace{2cm}
}

\begin{document}

\maketitle

\begin{abstract}
Facing the current debate on whether \acp{llm} attain near-human intelligence levels \citep{mitchell2023debate,bubeck2023sparks,kosinski2023theory,shiffrin2023probing,ullman2023large}, the current study introduces a benchmark for evaluating social intelligence, one of the most distinctive aspects of human cognition. We developed a comprehensive theoretical framework for social dynamics and introduced two evaluation tasks: \ac{ir} and \ac{iip}. Our approach also encompassed a computational model based on recursive Bayesian inference, adept at elucidating diverse human behavioral patterns. Extensive experiments and detailed analyses revealed that humans surpassed the latest GPT models in overall performance, zero-shot learning, one-shot generalization, and adaptability to multi-modalities. Notably, GPT models demonstrated social intelligence only at the most basic order (order = 0), in stark contrast to human social intelligence (order $\ge2$). Further examination indicated a propensity of \acp{llm} to rely on pattern recognition for shortcuts, casting doubt on their possession of authentic human-level social intelligence. Our codes, dataset, appendix and human data are released at \url{https://github.com/bigai-ai/Evaluate-n-Model-Social-Intelligence}.

\end{abstract}

\setstretch{0.95}

\section{Introduction}

The emergence of \acp{llm} has significantly influenced diverse fields, sparking debates about the potential emergence of \ac{agi}. Central to this discussion is whether \acp{llm} can match or surpass human intelligence \citep{bubeck2023sparks,openai2023gpt4,shiffrin2023probing}. Advocates suggest \acp{llm} exhibit key human intelligence markers, such as \ac{tom}, particularly in standard tasks like the false belief test \citep{kosinski2023theory}. However, critics point to a notable gap in \acp{llm}'s abilities, arguing they rely on superficial heuristics rather than deep \ac{tom} understanding and struggle with novel or slightly altered scenarios \citep{ullman2023large,sap2022neural,ma2023tomchallenges}. This is also evident in their handling of counterfactual and causal reasoning \citep{arkoudas2023gpt,webb2023emergent,binz2023using,ma2023brain,peng2023tong,collins2022structured}. Despite these revelations, a methodical, scientific framework for directly comparing machine and human intelligence is lacking.

\begin{figure}[t!]
    \centering
    \includegraphics[width=\linewidth]{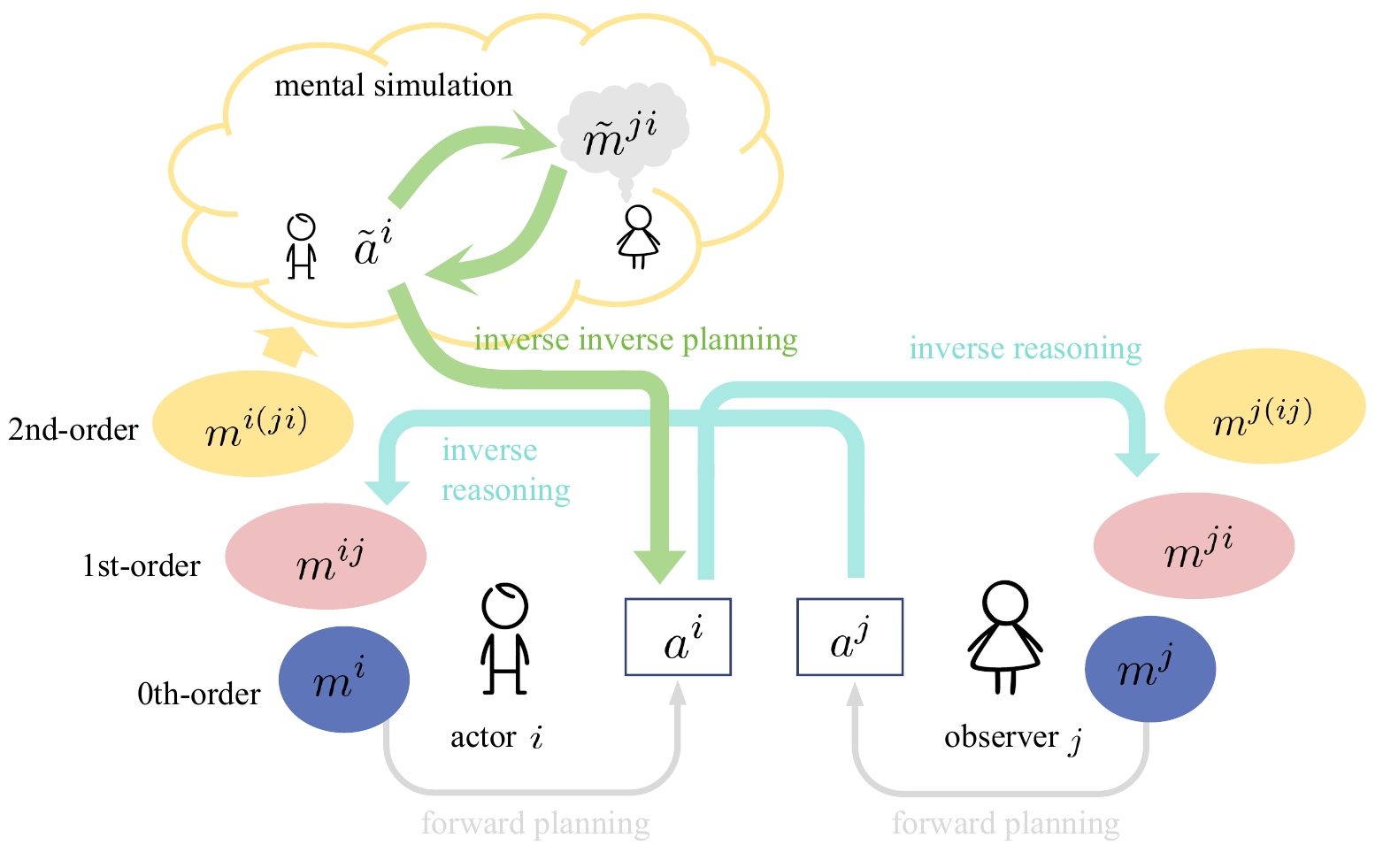}
    \caption{\textbf{A unified framework of social dynamics.} The foundational unit of human social interaction is exemplified by the actor $i$ and the observer $j$. This interaction is characterized by recursive mind reasoning, leading to the formation of a multi-layered cognitive architecture termed as ``\textbf{N Minds}'' \citep{fan2021learning}. This structure encompasses various levels of cognitive processing, including 0th-order minds, 1st-order minds, and 2nd-order minds. Our framework primarily concentrates on three critical mental operations: (i) \textbf{Forward Planning}, where actors strategize future actions based on current states; (ii) \textbf{Inverse Reasoning}, involving the observer's deduction of underlying actor motives from observed actions; and (iii) \textbf{Inverse Inverse Planning}, a higher-order cognitive process where the actor anticipates the observer's inferences and plans actions accordingly.}
    \label{fig:framework}
    \vspace{-0.22cm}
\end{figure}

Our research addresses this void by introducing a benchmark specifically designed for evaluating social intelligence, a key differentiator of human cognition from other primates \citep{fan2022artificial}. \citet{herrmann2007humans} revealed that while children and chimpanzees have similar cognitive abilities in physical tasks, children surpass both chimpanzees and orangutans in social tasks. Consequently, social intelligence emerges as a crucial metric for assessing whether \acp{llm} can match human cognitive abilities. We propose a comprehensive framework for social dynamics (\Cref{fig:framework}), focusing on key aspects of social interactions: social perception, \ac{tom} reasoning, and decision-making between two agents (the actor and the observer). The framework emphasizes three main processes: forward planning, inverse reasoning, and inverse inverse planning.

\begin{figure}[t!]
    \centering
    \includegraphics[width=\linewidth]{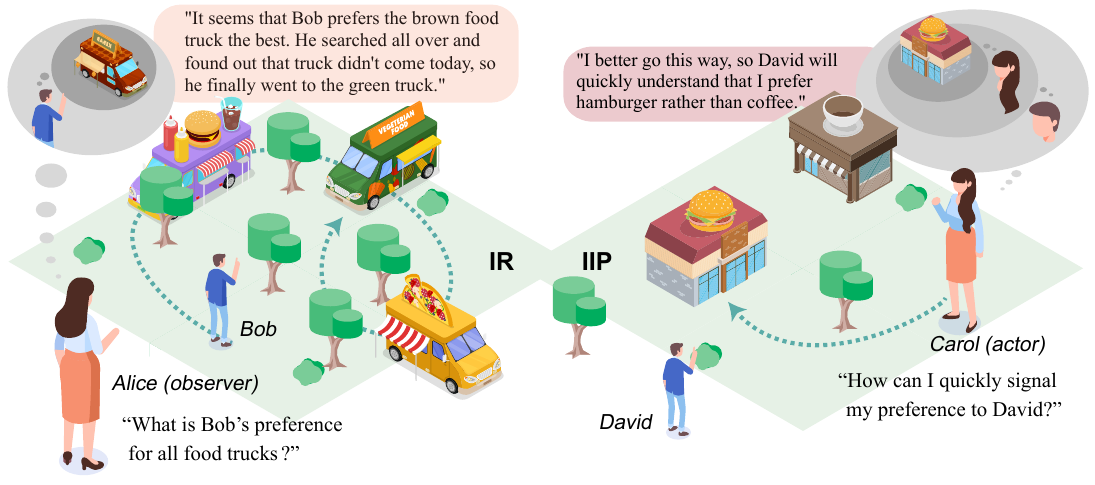}
    \caption{\textbf{Evaluation tasks: \ac{ir} (left) and \ac{iip} (right).} The \ac{ir} task involves observer Alice analyzing actor Bob's trajectory to deduce his preferred food truck. In the \ac{iip} task, actor Carol strategizes her route to efficiently convey her restaurant preference to observer David.}
    \label{fig:two_tasks}
    \vspace{-0.22cm}
\end{figure}

In our evaluation methodology, we introduce two key tasks: \acf{ir} and \acf{iip}. \citet{baker2017rational} studied the process of \ac{ir} in ``Food Truck'' task: inversely reason about human beliefs and preferences from their trajectories. Further, \citet{chandra2023acting} studied the \ac{iip} task: the actor plans actions to best convey desire. We extend them to more complicated versions (\Cref{fig:two_tasks}). Note that our selected tasks are designed to comprehensively encompass four cognitive dimensions: (i) rationality, (ii) perspective switching, (iii) counterfactual reasoning, (iv) and cognitive flexibility, thereby effectively evaluating social intelligence.

Additionally, we have developed a unified computational model that based on recursive Bayesian inference. This model interprets \ac{ir} as the observer's odd-order inference and \ac{iip} as the actor's even-order inference, providing a systematic approach to modeling intricate social interactions. Our model highlights the differences in preferences and decision-making processes between human cognition and machine approaches, delineating a clear distinction in how each comprehends social dynamics. Our extensive experimental studies and in-order analysis demonstrate that humans significantly surpass \acp{llm} in multiple aspects: overall performance, zero-shot learning, one-shot generalization, and adaptability to different modalities. We also find that the social intelligence demonstrated by \acp{llm} is only at the most rudimentary order (order = 0), in stark contrast to human social intelligence (order $\ge2$). And \acp{llm} are found to rely on pattern recognition for shortcuts, rather than possessing authentic human-level social intelligence. Our model closely aligns with human performance patterns, offering new perspectives in the ongoing discourse on human versus machine intelligence and contributing to the advancement of \ac{asi}.

In summary, by offering a nuanced benchmark for evaluating social intelligence, including a robust framework, representative tasks, an advanced computational model and benchmark experimental results of humans and machines (i.e., our model, LLMs), our work lays a foundational stone to bridge the gap, aspiring for a future where machines can more authentically replicate the human social intelligence intricacies.

\section{Related Work}


\paragraph{Cognitive Abilities for Social Intelligence}

\textit{Rationality} is considered a fundamental ability of an agent, denoting the capacity for optimal decision-making \citep{gergely1995taking,sodian2004infants}. Research indicates that infants as young as 12 months exhibit rationality in social contexts \citep{gergely1995taking}. \textit{Perspective switching} involves the capability to understand perspectives different from one's own, moving beyond a solely egocentric viewpoint \citep{underwood1982perspective,ackermann2012perspective}. By age 4, children begin to grasp that others may hold different perspectives \citep{ackermann2012perspective,borke1975piaget, baron1985does}. Notably, perspective switching is intricately linked to prosocial behavior \citep{ackermann2012perspective,stone2006theory,lemare1987perspective}, and its absence is a challenge in social interactions, particularly observed in individuals with autism \citep{underwood1982perspective}. \textit{Counterfactual reasoning} pertains to envisaging alternate outcomes based on different choices \citep{epstude2008functional,byrne2017counterfactual,beck2006children}, a skill that begins to develop in 2-year-olds and matures throughout childhood \citep{byrne2016counterfactual,byrne2017counterfactual,nyhout2019mature,rafetseder2010counterfactual}. There is significant evidence linking the development of this ability with \ac{tom} \citep{byrne2016counterfactual}. \textit{Cognitive flexibility} refers to the ability to adapt thoughts and actions in response to changing contexts \citep{dajani2015demystifying,ionescu2012exploring,barbey2021human,barbey2018network,barbey2013architecture,yakupov2022social}, and is fundamental to various cognitive capabilities, including task-switching in dual-task scenarios \citep{liu2016effect}.

\paragraph{Computational Models on Social Dynamics}

Social dynamics modeling often encompasses a dynamic feedback loop of actions, reactions, and cognitive processes between two agents \citep{kingsbury2020multi,schilbach2013toward}. Bayesian models, like Bayesian Inverse Planning and BToM, are employed to deduce others' mental states from observed behaviors \citep{baker2009action,baker2017rational}. \cite{chandra2023acting} extended these models to include ``inverse inverse planning,'' whereby agents strategically choose actions to shape audience perception. \cite{wang2020mathematical} developed mathematical models for 2-agent \ac{tom} of varying orders. In scenarios involving more than two agents, \cite{fan2021learning} introduced a structured mental representation termed ``N minds.''

\section{Evaluation Tasks}

In order to assess the social intelligence of both humans and \acp{llm}, we introduce two tasks, specifically \acf{ir} and \acf{iip}, adapted from \cite{baker2017rational} and \cite{chandra2023acting} respectively. The two representative tasks are designed to reflect four basic key cognitive dimensions, including rationality, perspective switching, counterfactual reasoning, and cognitive flexibility, as well as to encapsulate the three key mental processes inherent in human social interaction between an observer and an actor, especially ``Inverse Reasoning'' and ``Inverse Inverse Planning''. For illustrative details, refer to \Cref{fig:two_tasks}.

\subsection{Task 1: \texorpdfstring{\acf{ir}}{}}

\begin{figure}[t!]
    \centering
    
    \begin{subfigure}{0.33\linewidth}
        \includegraphics[width=\linewidth]{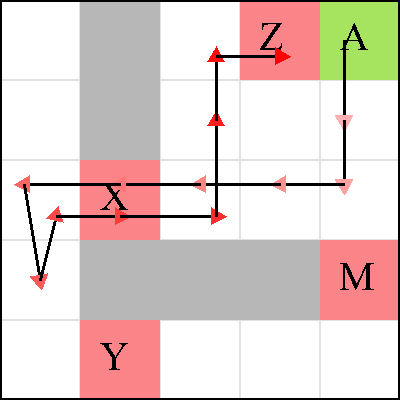}
        \caption{\ac{ir} task}
        \label{fig:stimuli_ir}
    \end{subfigure}%
    \hfill%
    \begin{subfigure}{0.33\linewidth}
        \includegraphics[width=\linewidth]{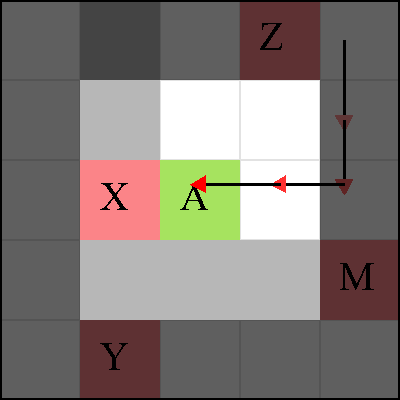}
        \caption{IR perception field}
        \label{fig:stimuli_ir_sight}
    \end{subfigure}%
    \hfill%
    \begin{subfigure}{0.33\linewidth}
        \includegraphics[width=\linewidth]{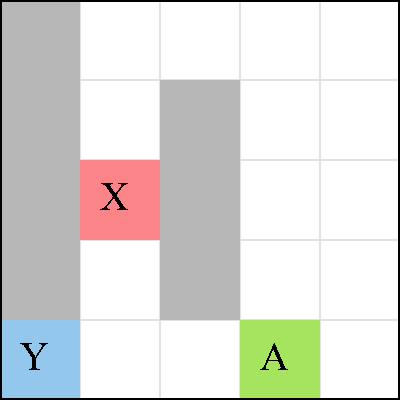}
        \caption{\ac{iip} task}
        \label{fig:stimuli_iip}
    \end{subfigure}%
    \\%
    \begin{subfigure}{0.2475\linewidth}
        \includegraphics[width=\linewidth]{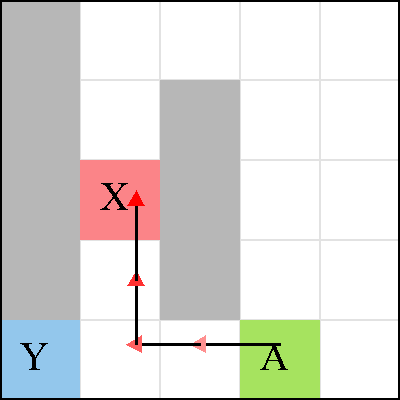}
        \captionsetup{justification=centering}
        \caption{\ac{iip} Route A \\ \nShort}
    \end{subfigure}%
    \hfill%
    \begin{subfigure}{0.2475\linewidth}
        \includegraphics[width=\linewidth]{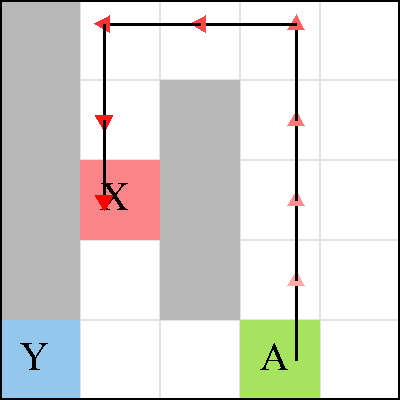}
        \captionsetup{justification=centering}
        \caption{\ac{iip} Route B \\ \nFar}
    \end{subfigure}%
    \hfill%
    \begin{subfigure}{0.2475\linewidth}
        \includegraphics[width=\linewidth]{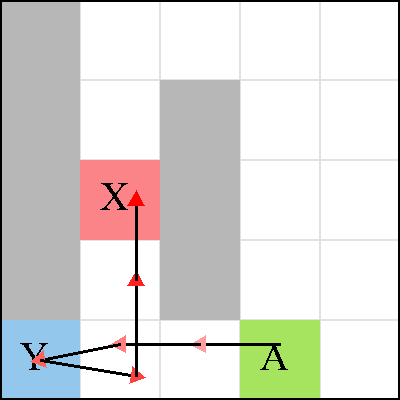}
        \captionsetup{justification=centering}
        \caption{\ac{iip} Route C \\ \nMisld}
    \end{subfigure}%
    \hfill%
    \begin{subfigure}{0.2475\linewidth}
        \includegraphics[width=\linewidth]{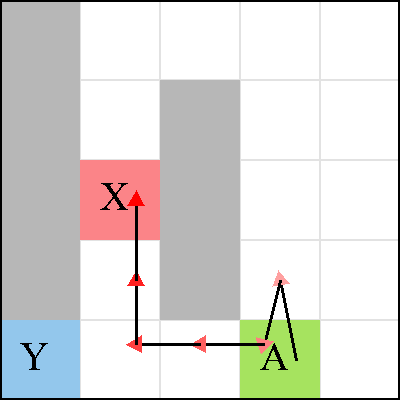}
        \captionsetup{justification=centering}
        \caption{\ac{iip} Route D \\ \nBest}
        \label{fig:iip_route_d}
    \end{subfigure}%
    \caption{\textbf{Input stimuli examples for both tasks.} (a) Scene layout and actor's trajectory in the \ac{ir} task; (b) Agent perception field in \ac{ir}; (c) Scene layout for the \ac{iip} task; (d)-(g) Four potential routes for the actor in the \ac{iip} task scenario. During testing, routes are randomly shuffled to ensure unbiased assessment.}
    \label{fig:input-stimuli}
\end{figure}

As depicted in \Cref{fig:stimuli_ir}, the \ac{ir} task takes place on a $5\times5$ grid campus with 4 parking slots, each highlighted in red. The setup includes 5 distinct food trucks, labeled $X$, $Y$, $Z$, $M$, and $N$. Every day, 4 of these trucks, say $X$, $Y$, $Z$, and $M$, are randomly allocated to the parking slots. Agent $A$ (in green) roams the campus with the aim of finding their most preferred food truck. Agent $A$'s preferences are \textbf{strict} (excluding equality, non-comparability, or cyclical preferences) and \textbf{stable} (consistent across time and location). The task operates in a partially observable setting, limiting Agent $A$'s vision to the immediate 8 cells and integrating occluding walls (in grey) to increase complexity. The task's objective is to analyze Agent $A$'s movement and infer their preference order for the food trucks, with some level of uncertainty in the answers being acceptable.

\begin{figure}[t!]
    \centering
    \begin{subfigure}{0.33\linewidth}
        \includegraphics[width=\linewidth]{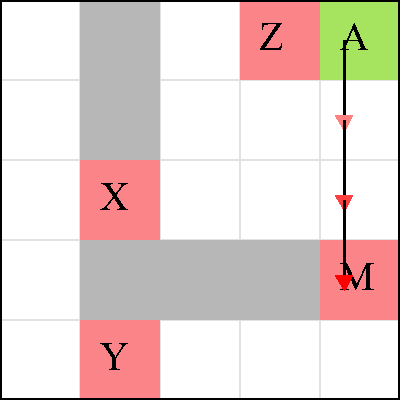}
        \caption{\nDirect}
    \end{subfigure}%
    \hfill%
    \begin{subfigure}{0.33\linewidth}
        \includegraphics[width=\linewidth]{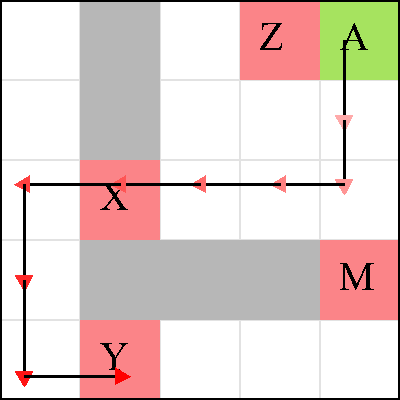}
        \caption{\nFinal}
    \end{subfigure}%
    \hfill%
    \begin{subfigure}{0.33\linewidth}
        \includegraphics[width=\linewidth]{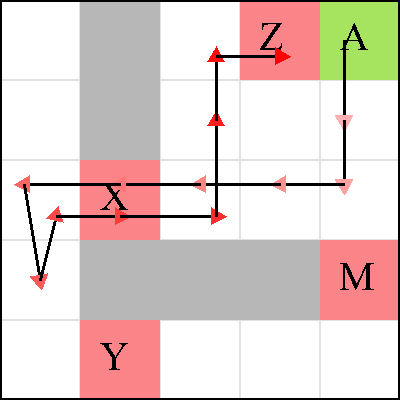}
        \caption{\nTurn}
    \end{subfigure}%
    \caption{\textbf{\ac{ir} task types.} (a) \nDirect: represented by $M>\{X,Y,Z,N\}$, indicates that $M$ is preferred over the others $X$, $Y$, $Z$, and $N$; (b) \nFinal: Characterized by $Y>\{X, Z, M\}$, suggests that $Y$ is chosen last among the visible options, leaving the preference for the absent $N$ as uncertain; (c) \nTurn: depicted as $N>Z>\{X,Y,M\}$, the actor revisits and chooses $Z$ after seeing all options, implying preference for $N$ over $Z$, and $Z$ over $X$, $Y$, and $M$.}
    \label{fig:ft_types}
\end{figure}

As detailed in \Cref{fig:ft_types}, each \ac{ir} problem is categorized into one of three distinct types, based on the actor's trajectory characteristics and the subsequent inference patterns.
\begin{itemize}[leftmargin=*,noitemsep,nolistsep]
    \item \textbf{\nDirect}: The actor concludes their route without exploring all food trucks. The selected one, which the actor stops at, is inferred to be the most preferred among all.
    \item \textbf{\nFinal}: The actor visits all available trucks, then select the last seen ($Y$ in \Cref{fig:ft_types}) directly. This choice suggests a preference order of $Y\!>\!\{X,Z,M\}$. However, the preference for the absent truck $N$ remains undetermined.
    \item \textbf{\nTurn}: After viewing all trucks, the actor retraces steps to a previously seen truck, such as $Z$. This behavior indicates a preference hierarchy where $N\!>\!Z\!>\!\{X,Y,M\}$.
\end{itemize}
These types include distinct strategies and decision-making processes, offering diverse insights into the actor's preference and cognitive mechanisms in social intelligence evaluation.

\subsection{Task 2: \texorpdfstring{\acf{iip}}{}}
\label{subsec:task2}

As illustrated in \Cref{fig:stimuli_iip}, the setting for the \ac{iip} task involves a $5\times5$ grid campus, two distinct restaurants $X$ and $Y$ (colored red and blue, respectively), and occluding walls (grey) on map. In this scenario, an agent $A$ (marked in green),  knowing the locations of both restaurants, prefers dining at $X$. The goal for $A$ is to demonstrate this preference to an observer $B$ through her movement route. She should express her preference on $X$ as early and unambiguous as possible, while also minimizing the travel length. It is assumed that $A$ is aware of $B$ being cooperative and capable of implicit understanding.

\begin{figure}[t!]
    \centering
    \begin{subfigure}{0.2475\linewidth}
        \includegraphics[width=\linewidth]{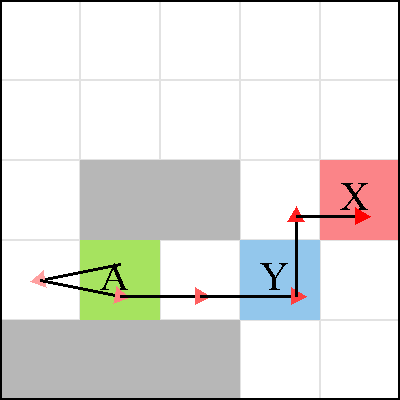}
        \caption{Type I}
    \end{subfigure}%
    \hfill%
    \begin{subfigure}{0.2475\linewidth}
        \includegraphics[width=\linewidth]{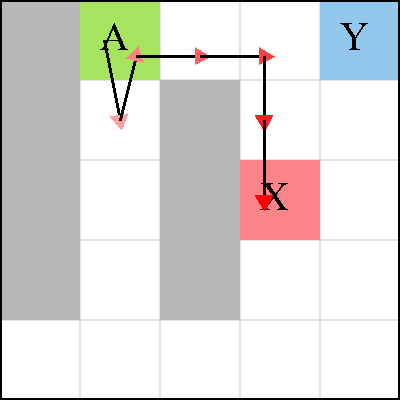}
        \caption{Type II}
    \end{subfigure}%
    \hfill%
    \begin{subfigure}{0.2475\linewidth}
        \includegraphics[width=\linewidth]{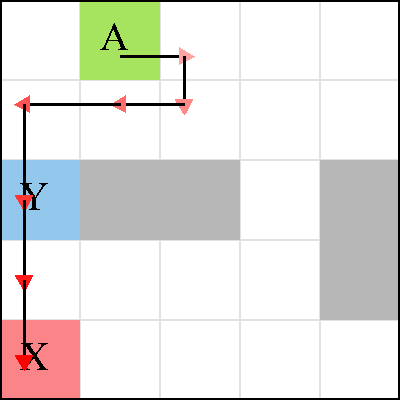}
        \caption{Type III}
    \end{subfigure}%
    \hfill%
    \begin{subfigure}{0.2475\linewidth}
        \includegraphics[width=\linewidth]{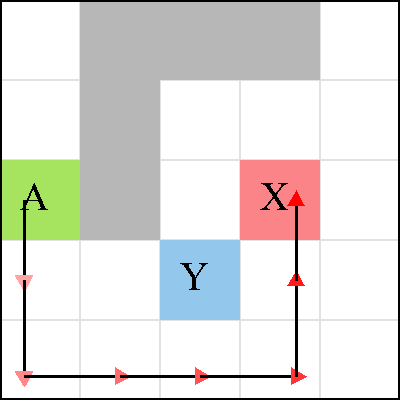}
        \caption{Type IV}
    \end{subfigure}%
    \caption{\textbf{\ac{iip} task types with \nBest{} routes.} (a) Type I: Cyclic route, revisiting a location and passing an alternative restaurant $Y$; (b) Type II: A cyclic route that does not entail passing through the vicinity of restaurant $Y$; (c) Type III: An acyclic route passing by the alternative restaurant $Y$; (d) Type IV: An acyclic route that avoids the vicinity of restaurant $Y$. Each type presents distinct problem patterns and difficulties for the actor to communicate their preference.}
    \label{fig:iip_types}
    \vspace{-0.2cm}
\end{figure}

Given the limitations of GPT-4 in route planning within grid environments \citep{borji2023categorical,bubeck2023sparks}, the \ac{iip} task (\Cref{fig:input-stimuli}(c)) is structured as a multiple-choice problem rather than route generation. The four candidate routes (\Cref{fig:input-stimuli}(d-g)) are generated by algorithms (see appendix). In terms of the four cognitive dimensions: (1) rationality, (2) perspective switching, (3) counterfactual reasoning and (4) cognitive flexibility, \nShort{} shoots at the shortest route to goal $X$, demonstrating (1) but no other dimensions; \nFar{} avoids restaurant $Y$ at the cost of route length, showing (1)(2)(3) but no (4); \nMisld{} signals ``I am not choosing $Y$'' by first arriving at $Y$ and then leaving $Y$ for $X$, with (1)(2)(3) but no (4); \nBest{} first uses a ``stepping away and back'' strategy to quickly signal ``my real goal is $X$ rather than the nearer $Y$'' at minimal route length cost, demonstrating (1)(2)(3)(4). Moreover, each \ac{iip} problem is classified into one of four types (Type I-IV) based on route \nBest{} as elaborated in \Cref{fig:iip_types}.

Two environments and datasets were constructed for \ac{ir} and \ac{iip} tasks, categorized by their respective problem types. The \ac{ir} dataset has 487 instances: 283 \nDirect, 86 \nFinal, and 118 \nTurn{} instances. The \ac{iip} dataset contains four types (I-IV), with 106, 135, 125, and 134 instances in each type, totaling 500 instances. Theoretically, the generation algorithms can generate all conceivable scenarios for both tasks.

\section{Computational Framework}

Our computational framework for social dynamics employs recursive Bayesian inference, effectively unifying the modeling of both \ac{ir} and \ac{iip} tasks. This framework's hierarchical structure stems from recursive social reasoning about mental states \citep{de2017negotiating,de2022higher}. Zero-order \ac{tom} represents an egocentric viewpoint without understanding others' mental states (\eg, ``I want a banana''). First-order \ac{tom} involves inferring others' mental states (\eg, ``I think he wants a banana''), while second-order \ac{tom} adds another layer of recursive inference (\eg, ``I think that he thinks that I want a banana''). This multi-layered approach to mental state inference provides a means to analyze various levels of social interaction and assess the progress in artificial social intelligence.

In our model, as depicted in \Cref{fig:framework}, we designate roles of actor $i$ and observer $j$. The term ``forward planning'' refers to actor $i$ devising action $a^i$ based on their 0th-order mind $m^i$. ``Inverse reasoning'' describes actor $i$ deducing their 1st-order mind $m^{ij}$—their perception of observer $j$'s mental state $m^j$—from $j$'s action $a^j$. ``Inverse inverse planning'' is a higher-order planning process incorporating inverse reasoning where actor $i$ simulates how observer $j$ might interpret $i$'s intent ($\tilde{m}^{ji}$) from action $\tilde{a}^i$ and selects action $a^i$ to effectively communicate a specific intent. This process involves the 2nd-order mind $m^{i(ji)}$ and exemplifies advanced human social intelligence. Although real-life social interactions are varied, we argue that our framework captures the core essence of most social dynamics.
\begin{figure}[!t]
    \centering
    \begin{subfigure}{0.2475\linewidth}
        \includegraphics[width=\linewidth, trim=0.5cm 0.5cm 0.5cm 0.5cm, clip]{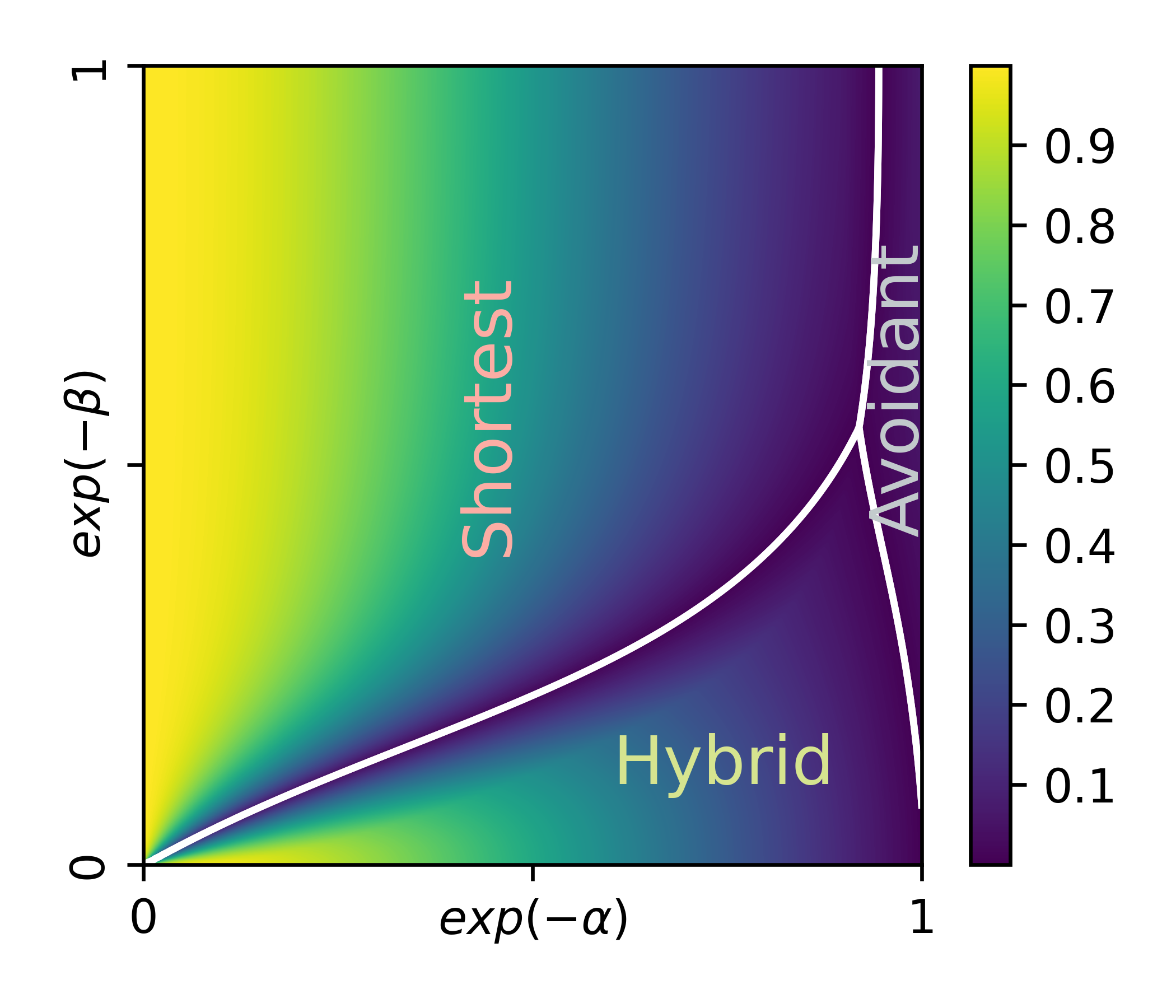}
        \caption{(0.3, 1)}
    \end{subfigure}%
    \begin{subfigure}{0.2475\linewidth}
        \includegraphics[width=\linewidth, trim=0.5cm 0.5cm 0.5cm 0.5cm, clip]{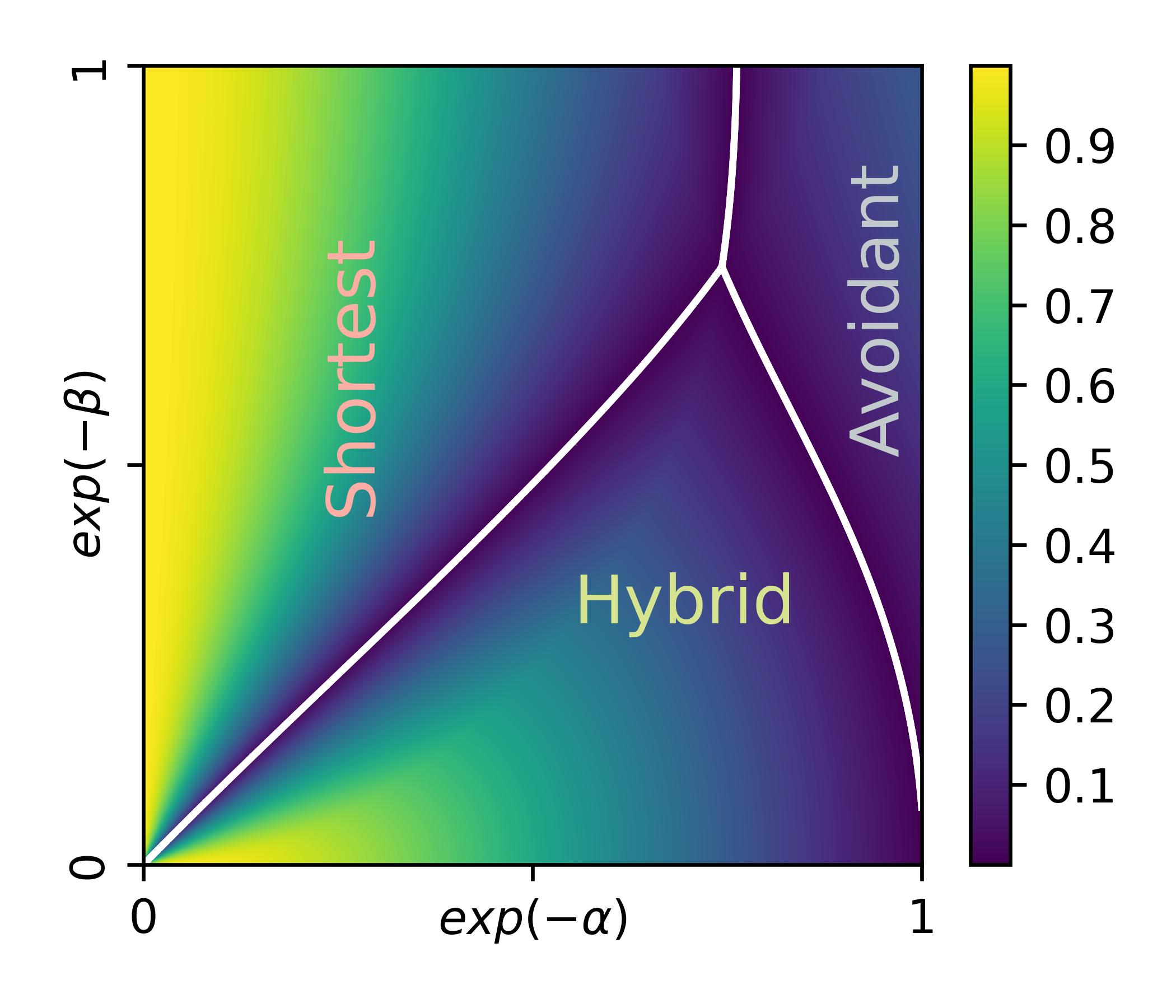}
        \caption{(0.99, 1)}
    \end{subfigure}%
    \begin{subfigure}{0.2475\linewidth}
        \includegraphics[width=\linewidth, trim=0.5cm 0.5cm 0.5cm 0.5cm, clip]{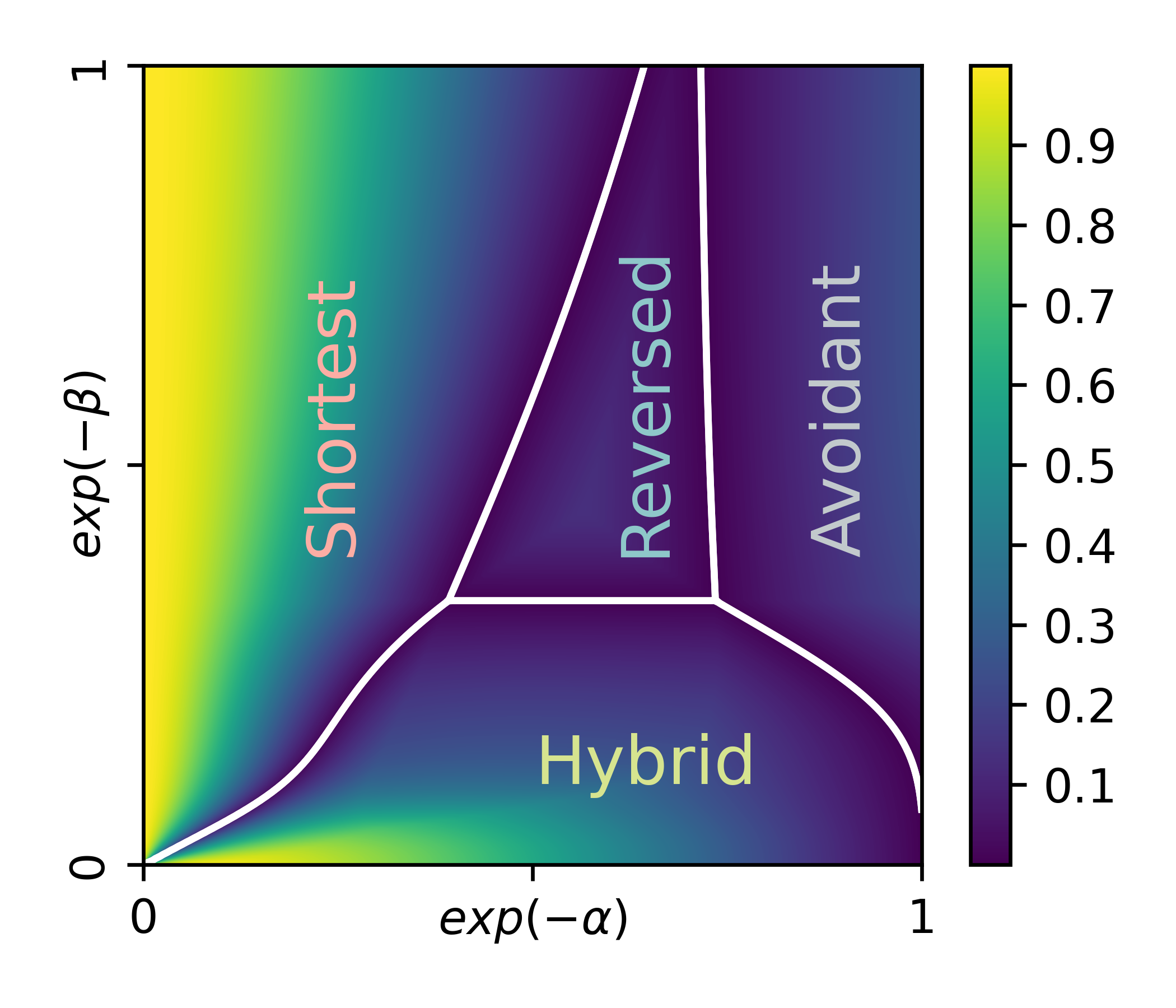}
        \caption{(0.3, 100)}
    \end{subfigure}
    \begin{subfigure}{0.2475\linewidth}
        \includegraphics[width=\linewidth, trim=0.5cm 0.5cm 0.5cm 0.5cm, clip]{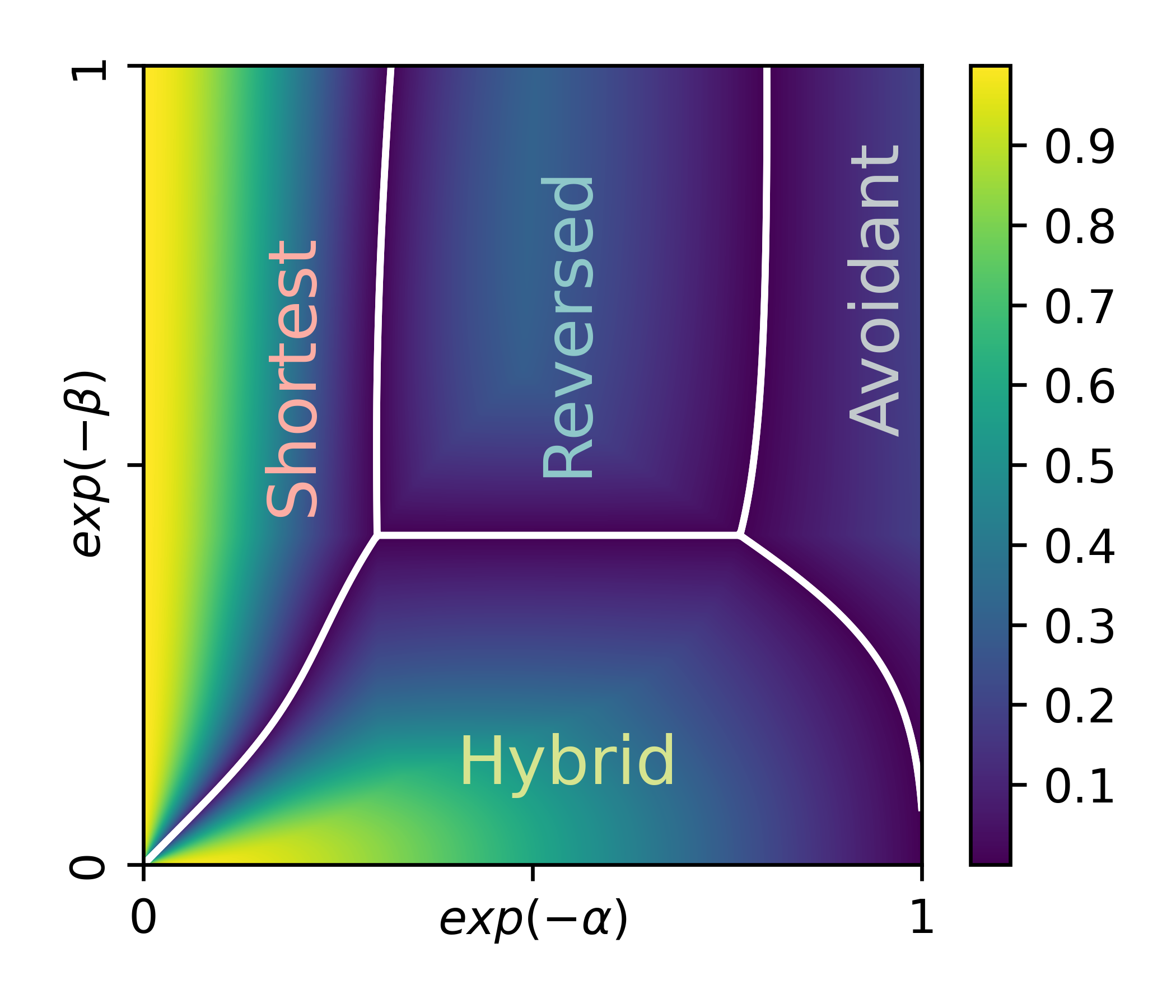}
        \caption{(0.99, 100)}
    \end{subfigure}%
    \caption{\textbf{Model predictions based on posterior probability over parameters $e^{-\alpha}$ and $e^{-\beta}$ on one example (\Cref{fig:input-stimuli}(c-g)).} The regions are designated according to the route types with the highest posterior. The color intensity within each region indicates the probability gap between the most likely and the second-most likely options, effectively visualizing the model's confidence in its predictions. Four figures are labeled by values of parameters $(exp(-\theta),\delta)$.}
    \label{fig:model_regions}
\end{figure}

\subsection{General Framework: Recursive Bayesian}

Consider an actor $i$, an observer $j$, and
the ``hypotheses set'' $\mathcal{H}$ as a set of elements $h$, each representing the information to be passed from $i$ to $j$, e.g.
actor $i$'s preference in the \ac{ir} task and their chosen target in the \ac{iip} task. 
Hypotheses are passed via routes $\gamma\in\Gamma$ in the
grid.
The recursive nature of our Bayesian framework emerges when \ac{tom} is integrated. This recursion reflects the agents' awareness of each other. Following \cite{yang2018optimal} and \cite{wang2020mathematical}, a two-agent recursive Bayesian system is defined at all orders, with each level depending on the inference made at the previous one, thus forming a sequential chain. We adopt the sequence that starts with actor $i$ acting independently at order 0, corresponding to a prior belief about $\Gamma$. Subsequently, observer $j$ makes inferences at order 1, followed by actor $i$ adjusting their behavior at order 2, and so on, as shown in Algorithm~\ref{alg:iter_bayesian}. This choice of sequence aligns with the notion of actor $i$ initially acting ``freely'' and the subsequent orders representing iterative inferences and reactions between the agents. 

Precisely, Algorithm~\ref{alg:iter_bayesian}
describes a belief drifting procedure in an \textbf{iterative} way. 
On even terms, the actor starts from the prior on routes $\Prob_p(\gamma)$. When considering the observer, the actor may choose to infer the observer's choice in mind, and to think $2k$ steps deeper resulting in a belief $\Prob^{2k}(\gamma|h)$. Similarly, the observer construct the odd terms $\Prob^{2k+1}(h|\gamma)$ using the same strategy.
The \textbf{recursive} explanation is, once an actor decides to think in order $2k$ (similar for observer in order 2k+1), the final Bayesian posterior $\Prob_i(\gamma|h)$ depends on the (2k-1)-st posterior $\Prob_{ij}(h|\gamma)$ on all possible $\gamma$ where $\Prob_{ij}$ means ``$j$ in $i$''. Then, $\Prob_{ij}(h|\gamma)$ depends on $\Prob_{iji}(\gamma|h)$ of order 2k-2, the ``($i$ in $j$) in $i$'' point of view, until order $0$ where $M$ and prior can be used.

\paragraph{\ac{ir} as Preference Inference: Odd-Order Inference}

In the \ac{ir} task, the hypothesis set $\mathcal{H}$ consists of full permutations of tuple $(X,Y,Z,M,N)$. \
A hypothesis $h=(Y\!>\!M\!>\!Z\!>\!N\!>\!X)$ has an array form, namely $h[0]=Y$, $h[1]=M$, etc. 
For the set $\Gamma$ of possible routes,  
we concentrate on the exploration order of trucks and the final decision. Among all such equivalent routes with same visiting order, we take the shortest one, thus route lengths are bounded by $(4+1)\times(5\times5)$\footnote{4+1 represents the 4 route segments in exploring the 4 trucks, plus a possible final segment to the chosen one; each route segment is no longer than the total amount of cells $5\times5$.},
and $\Gamma$ is finite. 
The task is for observer to infer the preference of actor, thus the result is an odd term in Algorithm~\ref{alg:iter_bayesian}.

\paragraph{\ac{iip} as Intentional Planning: Even-Order Inference}

The output of \ac{iip} is a posterior probability across four possible routes (an even term in Algorithm~\ref{alg:iter_bayesian}), denoted as $\Prob_{\aa}(\gamma|h)$. The hypothesis set $\mathcal{H}$ is limited to $\{X, Y\}$, and the route set $\Gamma$ consists of the options \{\nMisld, \nShort, \nFar, \nBest\}.


\begin{algorithm}
\caption{Iterative Bayesian Inference}
\label{alg:iter_bayesian}
\SetAlgoLined
\DontPrintSemicolon
\KwIn{Agents $i,j$, likelihood $M$, priors $\Prob_p(\gamma)$, $\Prob_p(h)$.}
\KwOut{Posteriors $\left(\Prob_p(\gamma),\Prob^1(h|\gamma),\Prob^2(\gamma|h),...\right)$.}
\BlankLine
\textbf{Initialize: $\Prob^0_i(\gamma|h)\propto M(\gamma,h)$}, $k=0$.\;
\For{$k=0$ \KwTo $\infty$}{
$\Prob^{2k+1}(h|\gamma):={\Prob^{2k}(\gamma|h)\Prob_p(h)}/{\Prob(\gamma)}$\;
$\Prob^{2k+2}(\gamma|h):={\Prob^{2k+1}(h|\gamma)\Prob_p(\gamma)}/{\Prob(h)}$\;
}
\KwRet{$\left(\Prob_p(\gamma),\Prob^1(h|\gamma),\Prob^2(\gamma|h),...\right)$.}
\end{algorithm}

\subsection{Detailed Construction for \texorpdfstring{\ac{ir}}{} and \texorpdfstring{\ac{iip}}{}}

Now we construct in detail the likelihoods and priors mentioned above, in a unified way, for both {\ac{ir}} and \ac{iip} to complete the model. 
As $\gamma$ is considered as a temporal signal sequence for $h$, the agent's sensitivity to signal urgency, cost and intensity are used as key factors for a unified construction.

We adopt a uniform distribution as the prior over $\mathcal{H}$, while the prior over $\Gamma$ is a Gibbs distribution of route lengths referring to the \textbf{total cost}: $\Prob(\gamma)\propto e^{-\alpha\cdot|\gamma|}$. Parameter $\alpha$ controls the sensitivity on cost. For the likelihood, let
\begin{equation}
    M(\gamma, h)\propto\sum\nolimits_{k=1}^{|\gamma|-1} \varphi(\gamma_{[0:k+1]},h)e^{-\beta k},
    \label{eq:likelihood_construction}
\end{equation}
where the route segment from $0$-th position to $t$-th ($\gamma_{[0:t+1]}$) is an element of the temporal signal at time $t$.
The parameter $\beta$ measures the urgency by the decay factor $e^{-\beta k}$ to the intensity of each route segment $\gamma_{[0:k+1]}$ represented by $\varphi$.

The function $\varphi$ represents the stimulus intensity, namely how likely a partial route indicates certain hypothesis. It is set to be a function to gain flexibility for both various tasks on the grid world and various styles of agents. A common setting could be $\varphi=\varphi_++\varphi_-$ the sum of accumulation effect $\varphi_+$ and elimination effect $\varphi_-$, both depending on task details.
Next, we provide constructions for \ac{ir} and \ac{iip}, respectively. 


\paragraph{Model for \ac{ir}}
Following the settings of \ac{ir}, cost sensitivity $\alpha$ is 0, according to the assumption that the actor looks for favourite on the map regardless of cost. The signal urgency $\beta$ is $-\infty$, since the whole route is available to the observer directly. For $\varphi$,
let $\mathcal{V}=\{X,Y,Z,M\}$ be set of all visible trucks,
$S(\gamma)\subset\mathcal{V}$ be set of trucks ever seen, $E(\gamma)\!\subset\!S(\gamma)$ be those seen but not chosen directly, and
$\varphi_+(\gamma_{[0:k+1]},h)=\mathbbm{1}_{\{\gamma_{[k]}\in\mathcal{V}\}\cap\{E(\gamma_{[0:k+1]})<4\}}(\gamma_{[0:k+1]})\cdot\mathbbm{1}_{\{h:\gamma_{[k]}=h[0]\}}(h)$ points out favourite,
$\varphi_-(\gamma_{[0:k+1]},h)=\mathbbm{1}_{\{\gamma_{[k]}\in\mathcal{V}\}\cap\{S(\gamma_{[0:k+1]})=4\}}(\gamma_{[0:k+1]})\cdot\mathbbm{1}_{\{h:h[0]=N,\gamma_{[k]}=h[1]\}}(h)$ considers what are not preferred.
Here $\mathbbm{1}_X(x)$ is the indicator function.
It can be shown that $h$'s with nonzero posterior match the analysis in previous section.


\paragraph{Model for \ac{iip}}

We set $\varphi=\varphi_++\varphi_-$, where $\varphi_+$ is modeled using a recursive coloring strategy (see appendix), influenced by a color-level amplification factor $\theta$, of form $\varphi_+(\gamma[0:k+1],h)=e^{-\theta\ell_h(\gamma[k])}$, while $\varphi_-$ represents a `negating' mechanism controlled by a leaving-target pulse $\delta$, i.e., $\varphi_-(\gamma[0:k+1],h)=\delta\mathbbm{1}_{\gamma[k-1]\in\mathcal{H}-\{h\}}(\gamma[0:k+1])$, signifying a firm rejection of the other target.
\Cref{fig:model_regions} demonstrates the model's behavior at order 2 for the \ac{iip} problem described in \Cref{fig:input-stimuli}, under varying parameters. It shows that appropriate settings of $\varphi$ (\eg, $e^{-\theta}=0.99$ and $\delta=100$) allow varying $\alpha$ and $\beta$ to generate all four choices, validating the model's reasonableness and expressiveness in \ac{iip} tasks.

\section{Experiments}

\paragraph{Our Model Implementation}

We developed our Bayesian model using Python, with PyTorch employed for gradient methods in MLE regression.

\paragraph{Human Participant Study}

Our study involved 75 participants who completed both the \ac{ir} and \ac{iip} tasks, presented in a randomized order. For the \ac{ir} task, each participant answered two questions from each of the three problem categories (\nDirect, \nFinal, \nTurn), and then answered two more questions following a \nTurn-type example. The \ac{iip} task consisted of a $4\times1+2$ format, where individuals first responded to one question from each of the four Types (I-IV) and then answered two more questions following a Type III example. Participants were randomly assigned to either a text-only or an image-enhanced multimodal version, labeled `human(text)' and `human(image)' respectively. The experiment concluded with a debriefing session for all participants.

\paragraph{\acp{llm} Evaluation}

We evaluated GPT-3.5-Turbo\footnote{\url{https://openai.com/blog/chatgpt}}, GPT-4-Turbo\footnote{\url{https://openai.com/blog/new-models-and-developer-products-announced-at-devday}}, and GPT-4 \citep{openai2023gpt4}, on the \ac{ir} and \ac{iip} tasks completely aligning to the text version of human study. Each problem of the entire problem database is tested in both zero-shot and one-shot settings, in a single round of conversation.

\section{Results and Analysis}

\begin{figure}[t!]
    \centering
    \begin{subfigure}{0.47\linewidth}
        \includegraphics[width=\linewidth, trim=0.1cm 0.2cm 0.4cm 0.2cm, clip]{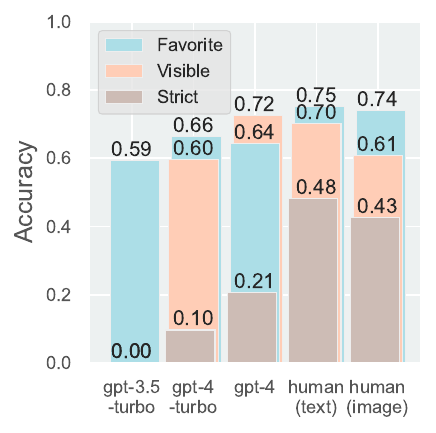}
        \caption{Overall Zero-Shot}
    \end{subfigure}%
    \hfill%
    \begin{subfigure}{0.47\linewidth}
        \includegraphics[width=\linewidth, trim=0.1cm 0.2cm 0.4cm 0.2cm, clip]{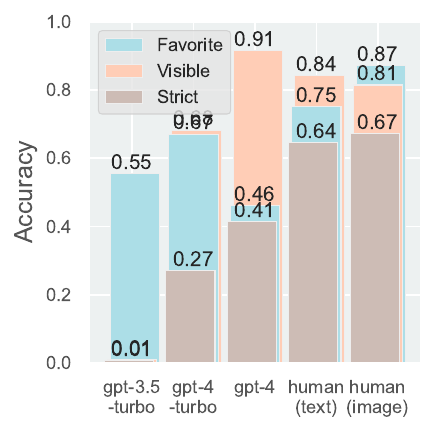}
        \caption{Overall One-Shot}
    \end{subfigure}
    \\%
    \begin{subfigure}{0.47\linewidth}
        \includegraphics[width=\linewidth, trim=0.1cm 0.2cm 0.4cm 0.2cm, clip]{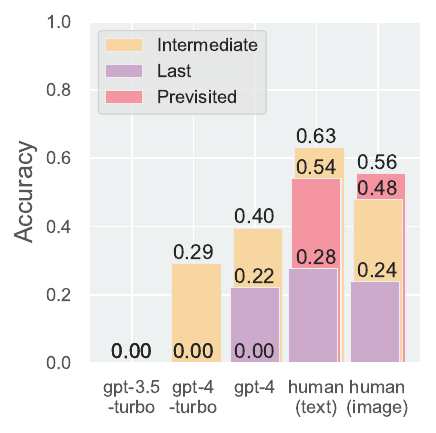}
        \caption{Type-specific Zero-Shot}
    \end{subfigure}%
    \hfill%
    \begin{subfigure}{0.47\linewidth}
        \includegraphics[width=\linewidth, trim=0.1cm 0.2cm 0.4cm 0.2cm, clip]{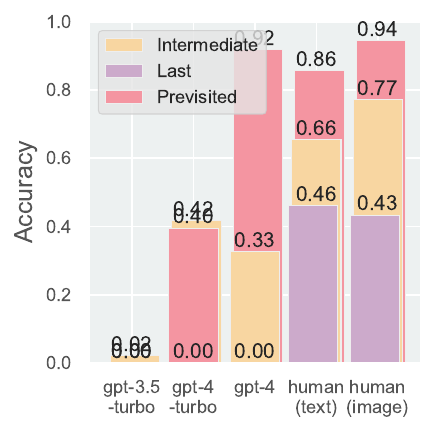}
        \caption{Type-specific One-Shot}
    \end{subfigure}%
    \caption{\textbf{Accuracy on the \ac{ir} Task.} In (a) and (b), ``Favorite'' assesses accuracy for the top preference only, ``Visible'' for the preference order among $\{X,Y,Z,M\}$, and ``Strict'' for the entire preference order. In (b) and (d), we uniformly use a \nTurn{} type case as the one-shot learning example. In (c) and (d), accuracies are evaluated solely based on the ``Strict'' criterion.}
    \label{fig:ft_results}
    \vspace{-0.2cm}
\end{figure}

\begin{figure}[t!]
    \centering
    \begin{subfigure}{0.49\linewidth}
        \includegraphics[width=\linewidth, trim=0.1cm 0.2cm 0.3cm 0.1cm, clip]{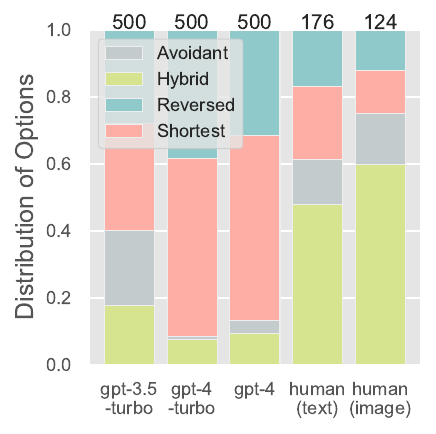}
        \caption{Overall Zero-Shot}
    \end{subfigure}%
    \hfill%
    \begin{subfigure}{0.49\linewidth}
        \includegraphics[width=\linewidth, trim=0.1cm 0.2cm 0.3cm 0.1cm, clip]{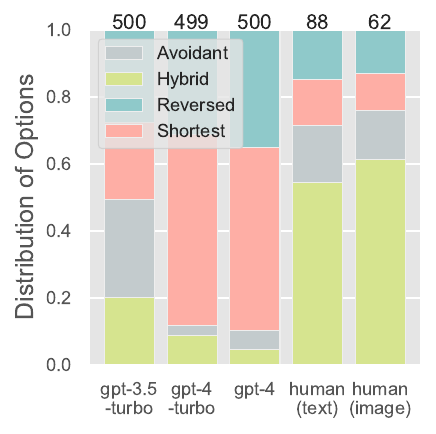}
        \caption{Overall One-Shot}
    \end{subfigure}
    \\%
    \begin{subfigure}{0.49\linewidth}
        \includegraphics[width=\linewidth, trim=0.1cm 0.2cm 0.3cm 0.1cm, clip]{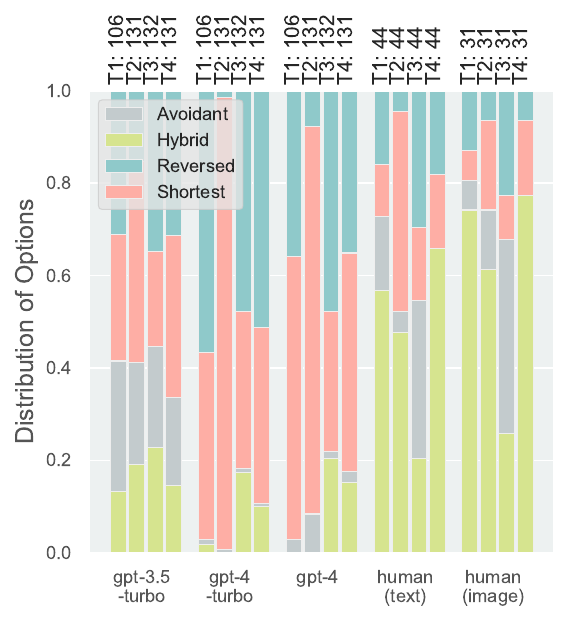}
        \caption{Type-specific Zero-Shot}
    \end{subfigure}%
    \hfill%
    \begin{subfigure}{0.49\linewidth}
        \includegraphics[width=\linewidth, trim=0.1cm 0.2cm 0.3cm 0.1cm, clip]{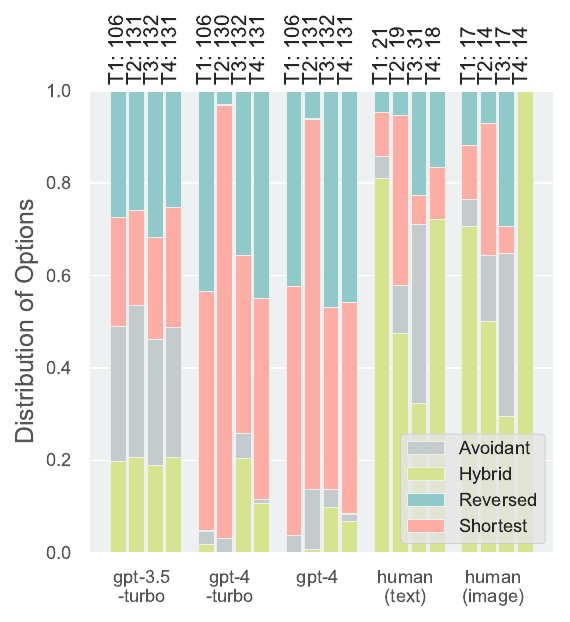}
        \caption{Type-specific One-Shot}
    \end{subfigure}%
    \caption{\textbf{Distribution of Options in \ac{iip}.} The numerical values at top of each bar represent the respective test counts. In (b) and (d), we uniformly use a Type III case as the one-shot learning example.}
    \label{fig:iip_stats}
    \vspace{-0.2cm}
\end{figure}

The evaluation of the \ac{ir} task shows the accuracy under various criteria and across different problem types for all participant groups, as illustrated in \Cref{fig:ft_results}. Similarly, for the \ac{iip} task, \Cref{fig:iip_stats} presents the statistical distributions under zero-shot or one-shot settings as well as across ``overall'' (aggregating four types) and type-specific settings.



Results indicate GPT-3.5-Turbo's inability to grasp the tasks. GPT-4 variants exhibited a pronounced tendency to select \nShort{} in \ac{iip}. In zero-shot settings, the GPT series displayed constrained counterfactual reasoning abilities, struggling with the concept of an unseen `$N$' (as shown in `Visible' and `Strict' categories in \Cref{fig:ft_results}(a) and the \nTurn{} category in \Cref{fig:ft_results}(c)). This suggests that GPT-4's capability in active \ac{tom} may not extend beyond a superficial level. Furthermore, GPT-4's one-shot enhancements were observed only in \ac{ir} tasks matching the example's type and were virtually absent in \ac{iip} tasks. This pattern implies that GPT-4's performance may not stem from an in-depth \ac{tom} understanding. The analyses from \Cref{fig:iip_stats} show that human participants generally exhibited \ac{tom} abilities at order $\ge2$, i.e., preferring route \nBest{} and showing all four cognitive dimensions (see \nameref{subsec:task2}). Following the one-shot example, human performance improved across all \ac{ir} categories, and there was a notable decline in the choice of \nShort{} options in \ac{iip}. This indicates a significant learning and generalization capability in social cognition tasks among humans.

\paragraph{Text vs. Image: Multimodal Capabilities}


Our focused case study (see appendix) indicates that image inputs to GPT-4V\footnote{\url{https://cdn.openai.com/papers/GPTV_System_Card.pdf}.} fails to  significantly enhance GPT capabilities, which still exhibit a considerable gap compared to human performance.

\paragraph{\ac{iip} Preference Regression}

\begin{figure}[t!]
    \centering
    \begin{subfigure}{0.49\linewidth}
        \includegraphics[width=\linewidth, trim=0 10 0 10, clip]{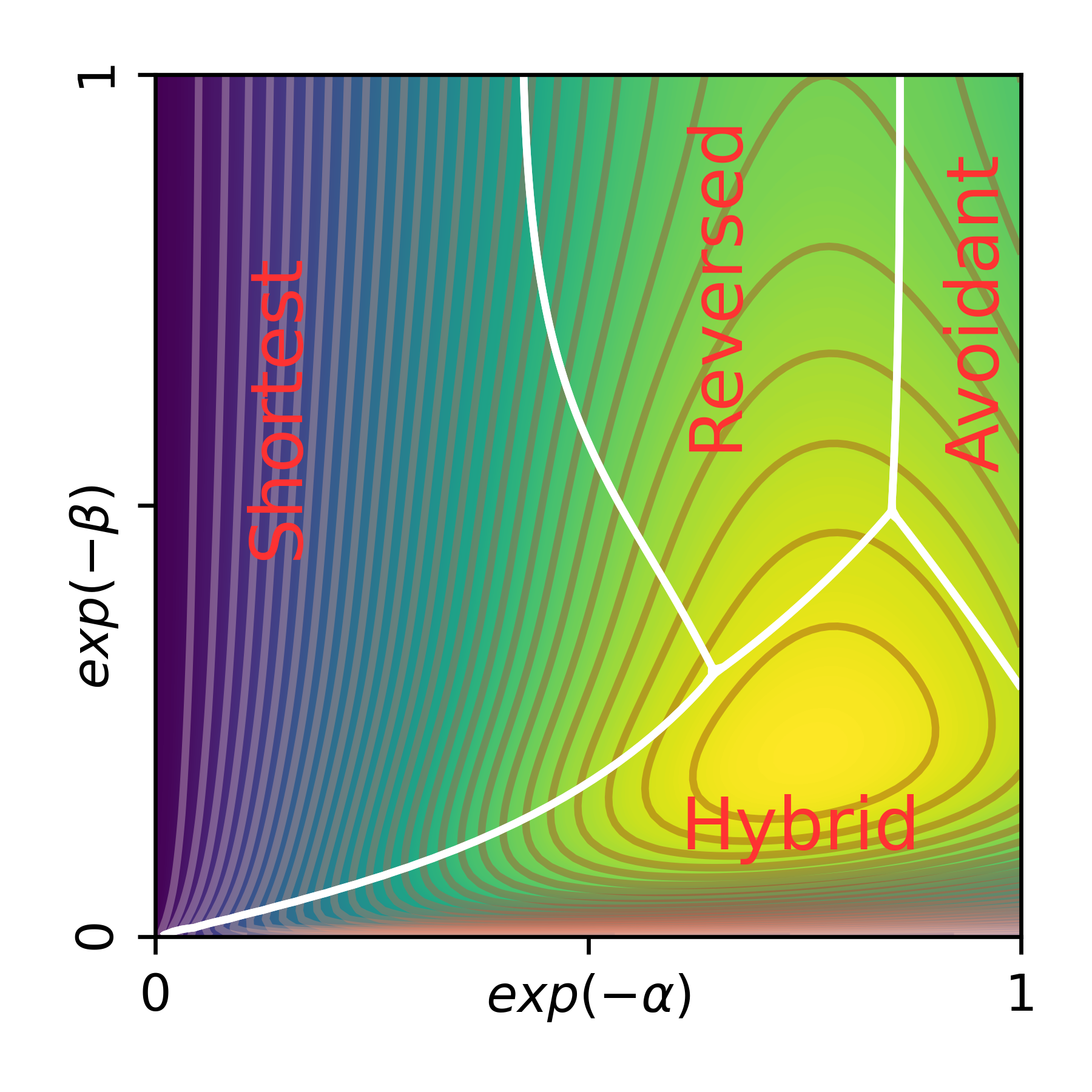}
        \caption{Human}
    \end{subfigure}%
    \hfill%
    \begin{subfigure}{0.49\linewidth}
        \includegraphics[width=\linewidth, trim=0 10 0 10, clip]{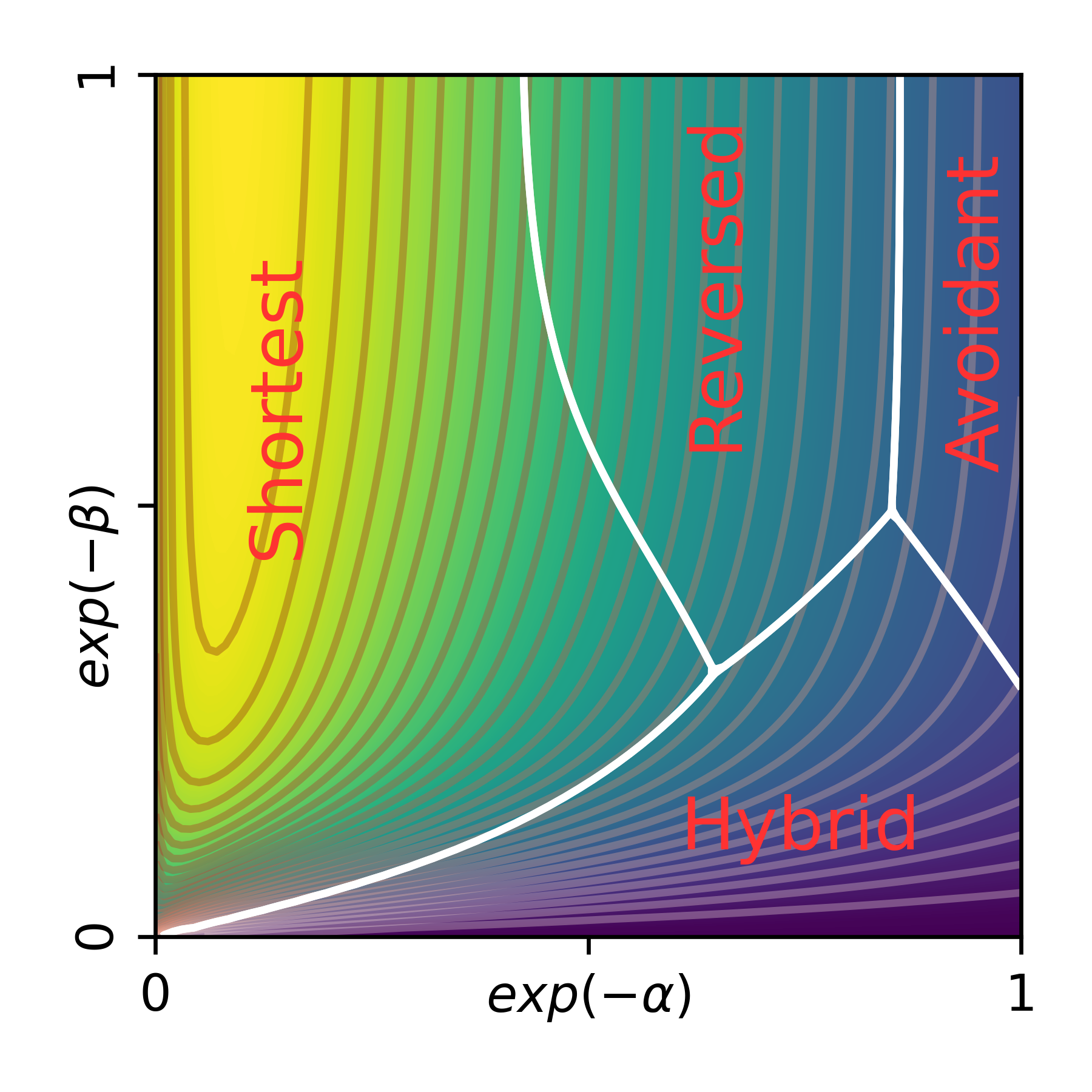}
        \caption{GPT-4}
    \end{subfigure}
    \hfill%
    \begin{subfigure}{0.49\linewidth}
        \includegraphics[width=\linewidth, trim=0 10 0 10, clip]{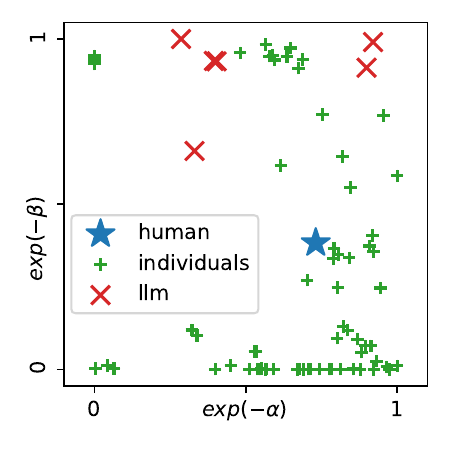}
        \caption{$\alpha$-$\beta$ plane}
    \end{subfigure}%
    \hfill%
    \begin{subfigure}{0.49\linewidth}
        \includegraphics[width=\linewidth, trim=0 10 0 10, clip]{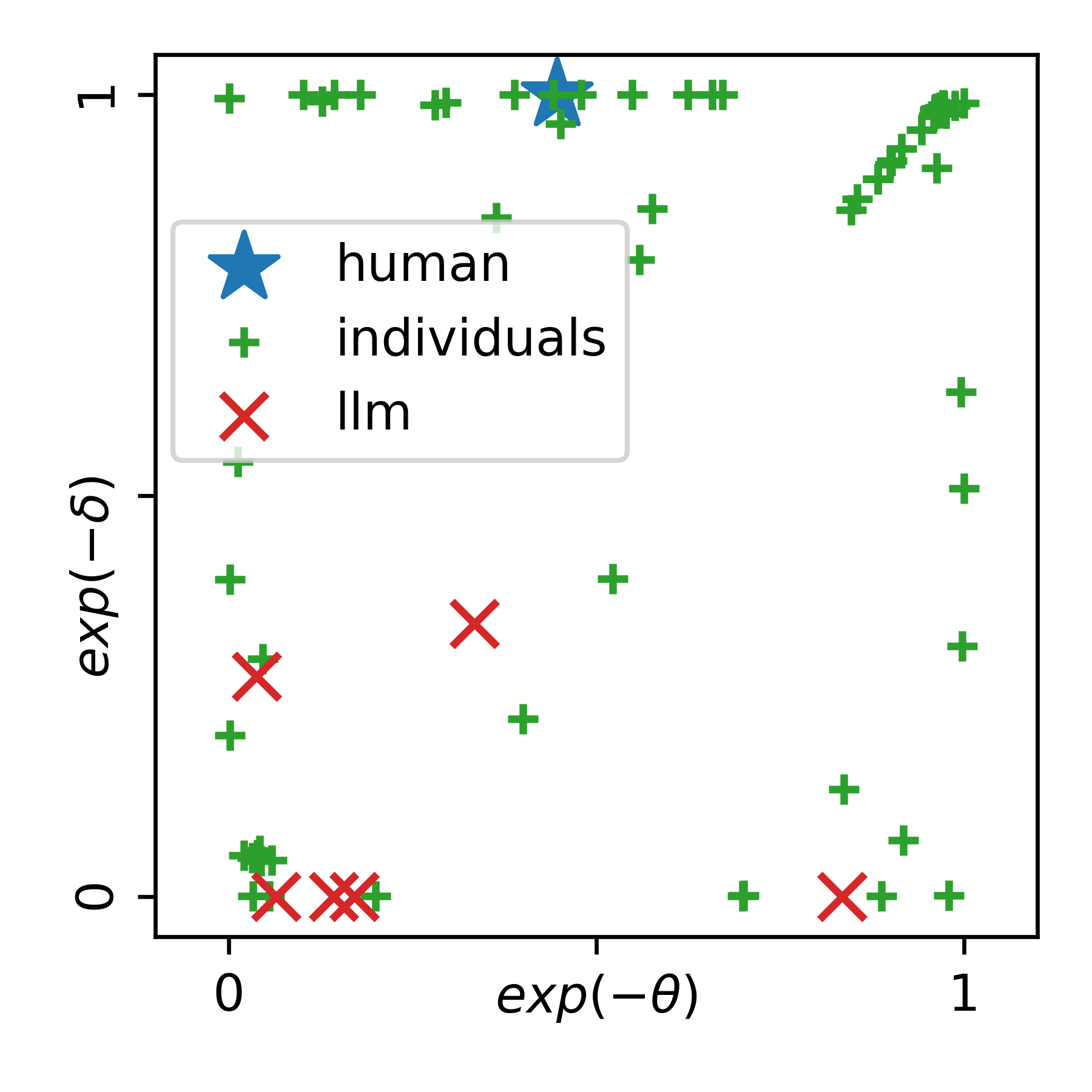}
        \caption{$\theta$-$\delta$ plane}
    \end{subfigure}%
    \caption{\textbf{\ac{iip} modeling results.} (a-b) Likelihood landscapes in the $\alpha$-$\beta$ dimension ($e^{-\theta}=0.99$, $\delta=100$), comparing ``human average'' with ``GPT-4''; region boundaries and labels are calculated as in \Cref{fig:model_regions} on the \textbf{whole dataset}. (c-d) Regression for human average, \acp{llm} and individual humans, mapped onto two planes respectively.}
    \label{fig:regression}
\end{figure}

Applying MLE to \ac{iip} test data (including \acp{llm}, individual human subjects, and the human average) allows for the regression of parameters within our model framework (\Cref{eq:likelihood_construction}). Setting $e^{-\theta}=0.99, \delta=100$, we plot the likelihoods of $\alpha,\beta$ for both humans and GPT-4 models in \Cref{fig:regression} (a-b), with regression outcomes depicted in (c-d). The patterns between humans and \acp{llm} diverge significantly. Additionally, referencing \Cref{fig:model_regions} reveals that despite considerable variability among individuals, a majority of humans tend to prefer the \nBest{} option. Conversely, GPT-4 displays a mixed preference for \nShort{} and \nMisld{}, aligning with the observed statistics in \Cref{fig:iip_stats}.

\paragraph{Shortcuts in \ac{ir} and \ac{iip} Tasks}

We investigated whether \acp{llm} rely on pattern recognition (shortcuts) rather than genuine social intelligence in tackling \ac{ir} and \ac{iip} tasks. In the \ac{ir} task, using the grid environment layout and trajectory as input, we post-finetuned a small model T5 \citep{raffel2020exploring} on specific task types and evaluated the \ac{ir} task accuracy under ``strict'' cirterion across all types. As demonstrated in \Cref{tab:ft_shortcut}, when trained on all task types, T5 can achieve high task accuracy across all task types; but, its performance on IR task significantly drops to $0$ when certain task types are absent in training, unlike humans who can achieve high task accuracy in ``zero-shot'' and ``one-shot'' setting (\Cref{fig:ft_results}(c)(d)). For the \ac{iip} task, we perform two route classification tasks. Firstly, we use only routes and no task contexts as input and route types as labels for the overall classification test on T5; as shown in \Cref{tab:iip_shortcut_basic}, T5 achieves very high performances, indicating clear pattern differences among different types of routes, which might be a shortcut for machines to memorize the better answer without analyzing the specific \ac{iip} task. Secondly, we perform a task-type-specific version of route type classification test on T5, using the grid environment layout and four candidate routes as input, and the corresponding four-route-type-in-order sequence as the label. As demonstrated in \Cref{tab:Iiip_shortcut}, there are also significant performance drops when T5 meets certain task type for the first time in testing without any data of that type in training. These shortcut experiments illustrate that, even if model finetuning on our data achieves high performance in the two tasks, it is insufficient to conclude that the model possesses strong social intelligence capabilities-it may only memorize the surface pattern shortcuts without deep reasoning; and unlike humans, it can not transfer its ability to unseen cases. Thereby, we should pay more attention to model's zero-shot and few-shot learning abilities.

\begin{table}[ht!]
    \centering
    \small
    \caption{\textbf{\ac{ir} shortcuts analysis.} We use \ac{ir} task accuracy (\%) under the ``strict'' criterion as the metric.}
    \label{tab:ft_shortcut}
    \resizebox{\linewidth}{!}{%
        \begin{tabular}{lccccc}
        \toprule
        & \nDirect & \nFinal & \nTurn & Avg \\
        \midrule
        Overall & 92.57 & 97.14 & 100.00 & 96.60 \\
        w/o \nFinal & 81.27 & 0.00 & 95.76 & 59.00 \\ 
        w/o \nDirect/\nFinal & 0.00 & 0.00 & 100.00 & 33.33\\ 
        w/o \nFinal/\nTurn & 100.0 & 0.00 & 0.00 & 33.33 \\ \bottomrule
        \end{tabular}%
    }%
\end{table}

\begin{table}[ht!]
    \centering
    \small
    \caption{\textbf{\ac{iip} shortcuts analysis for basic options.} We use route type classification accuracy (\%) as the metric.}
    \label{tab:iip_shortcut_basic}
        \begin{tabular}{lccccc}
        \toprule
        & \nMisld & \nShort & \nFar & \nBest  & Avg \\
        \midrule
        Overall & 99.4  & 95.2 & 91.0  & 94.2 & 94.9 \\
        \bottomrule
        \end{tabular}%
\end{table}

\begin{table}[ht!]
    \centering
    \caption{\textbf{\ac{iip} shortcuts analysis.} We use route type classification accuracy (\%) as the metric.}
    \label{tab:Iiip_shortcut}
    \resizebox{\linewidth}{!}{%
        \begin{tabular}{lccccc}
            \toprule
            & Type I & Type II & Type III & Type IV  & Avg\\
            \midrule
            Overall & 98.11 & 100.00 & 91.66  & 79.39 & 92.00\\ 
            w/o Type I & 94.33 & 98.47 & 94.69 & 90.07 & 94.40\\ 
            w/o Type II & 99.05 & \textbf{66.41}(-33.59) & 90.90 & 82.44 & 84.00\\ 
            w/o Type III & 100.00 & 99.23 & \textbf{52.27}(-39.39)  & 83.96 & 83.00\\ 
            w/o Type IV & 100.00 & 100.00 & 96.21 & \textbf{35.87}(-43.52) & 82.20\\ 
            w/o Type I,II & \textbf{65.09}(-33.02) & \textbf{13.74}(-86.26) & 87.88 & 81.68 & \textbf{62.00}\\ 
            w/o Type III,IV & 100.00 & 100.00 & \textbf{36.36}(-55.3) & \textbf{4.58}(-74.81) & \textbf{58.20}\\
            \bottomrule
        \end{tabular}%
    }%
\end{table}

\section{Conclusion}

We introduced a comprehensive benchmark for evaluating social intelligence, comprising a unified computational framework, representative tasks, and evaluation criteria. Our results demonstrate a marked superiority of humans over \acp{llm} in social intelligence tasks. We hope that our study contributes valuable information towards the advancement of \ac{asi}.

\paragraph{Acknowledgement}

This work is supported in part by the National Science and Technology Major Project (2022ZD0114900) and the Beijing Nova Program.

\bibliographystyle{apacite}
\renewcommand\bibliographytypesize{\small}
\setlength{\bibleftmargin}{.125in}
\setlength{\bibindent}{-\bibleftmargin}
\bibliography{reference_header,reference}

\begin{thebibliography}{}

\bibitem [\protect \citeauthoryear {%
Ackermann%
}{%
Ackermann%
}{%
{\protect \APACyear {2012}}%
}]{%
ackermann2012perspective}
\APACinsertmetastar {%
ackermann2012perspective}%
\begin{APACrefauthors}%
Ackermann, E.%
\end{APACrefauthors}%
\unskip\
\newblock
\APACrefYearMonthDay{2012}{}{}.
\newblock
{\BBOQ}\APACrefatitle {Perspective-taking and object construction: Two keys to learning} {Perspective-taking and object construction: Two keys to learning}.{\BBCQ}
\newblock
\BIn{} \APACrefbtitle {Constructionism in Practice} {Constructionism in practice}\ (\BPGS\ 25--35).
\newblock
\APACaddressPublisher{}{Routledge}.
\PrintBackRefs{\CurrentBib}

\bibitem [\protect \citeauthoryear {%
Arkoudas%
}{%
Arkoudas%
}{%
{\protect \APACyear {2023}}%
}]{%
arkoudas2023gpt}
\APACinsertmetastar {%
arkoudas2023gpt}%
\begin{APACrefauthors}%
Arkoudas, K.%
\end{APACrefauthors}%
\unskip\
\newblock
\APACrefYearMonthDay{2023}{}{}.
\newblock
{\BBOQ}\APACrefatitle {GPT-4 Can't Reason} {Gpt-4 can't reason}.{\BBCQ}
\newblock
\APACjournalVolNumPages{arXiv preprint arXiv:2308.03762}{}{}{}.
\PrintBackRefs{\CurrentBib}

\bibitem [\protect \citeauthoryear {%
Baker%
, Jara-Ettinger%
, Saxe%
\BCBL {}\ \BBA {} Tenenbaum%
}{%
Baker%
\ \protect \BOthers {.}}{%
{\protect \APACyear {2017}}%
}]{%
baker2017rational}
\APACinsertmetastar {%
baker2017rational}%
\begin{APACrefauthors}%
Baker, C\BPBI L.%
, Jara-Ettinger, J.%
, Saxe, R.%
\BCBL {}\ \BBA {} Tenenbaum, J\BPBI B.%
\end{APACrefauthors}%
\unskip\
\newblock
\APACrefYearMonthDay{2017}{}{}.
\newblock
{\BBOQ}\APACrefatitle {Rational quantitative attribution of beliefs, desires and percepts in human mentalizing} {Rational quantitative attribution of beliefs, desires and percepts in human mentalizing}.{\BBCQ}
\newblock
\APACjournalVolNumPages{Nature Human Behaviour}{1}{4}{0064}.
\PrintBackRefs{\CurrentBib}

\bibitem [\protect \citeauthoryear {%
Baker%
, Saxe%
\BCBL {}\ \BBA {} Tenenbaum%
}{%
Baker%
\ \protect \BOthers {.}}{%
{\protect \APACyear {2009}}%
}]{%
baker2009action}
\APACinsertmetastar {%
baker2009action}%
\begin{APACrefauthors}%
Baker, C\BPBI L.%
, Saxe, R.%
\BCBL {}\ \BBA {} Tenenbaum, J\BPBI B.%
\end{APACrefauthors}%
\unskip\
\newblock
\APACrefYearMonthDay{2009}{}{}.
\newblock
{\BBOQ}\APACrefatitle {Action understanding as inverse planning} {Action understanding as inverse planning}.{\BBCQ}
\newblock
\APACjournalVolNumPages{Cognition}{113}{3}{329--349}.
\PrintBackRefs{\CurrentBib}

\bibitem [\protect \citeauthoryear {%
Barbey%
}{%
Barbey%
}{%
{\protect \APACyear {2018}}%
}]{%
barbey2018network}
\APACinsertmetastar {%
barbey2018network}%
\begin{APACrefauthors}%
Barbey, A\BPBI K.%
\end{APACrefauthors}%
\unskip\
\newblock
\APACrefYearMonthDay{2018}{}{}.
\newblock
{\BBOQ}\APACrefatitle {Network neuroscience theory of human intelligence} {Network neuroscience theory of human intelligence}.{\BBCQ}
\newblock
\APACjournalVolNumPages{Trends in Cognitive Sciences}{22}{1}{8--20}.
\PrintBackRefs{\CurrentBib}

\bibitem [\protect \citeauthoryear {%
Barbey%
}{%
Barbey%
}{%
{\protect \APACyear {2021}}%
}]{%
barbey2021human}
\APACinsertmetastar {%
barbey2021human}%
\begin{APACrefauthors}%
Barbey, A\BPBI K.%
\end{APACrefauthors}%
\unskip\
\newblock
\APACrefYearMonthDay{2021}{}{}.
\newblock
{\BBOQ}\APACrefatitle {Human Intelligence and Network Neuroscience} {Human intelligence and network neuroscience}.{\BBCQ}
\newblock
\BIn{} \APACrefbtitle {The Cambridge Handbook of Intelligence and Cognitive Neuroscience} {The cambridge handbook of intelligence and cognitive neuroscience}\ (\BPGS\ 102--122).
\newblock
\APACaddressPublisher{}{Cambridge University Press}.
\PrintBackRefs{\CurrentBib}

\bibitem [\protect \citeauthoryear {%
Barbey%
, Colom%
\BCBL {}\ \BBA {} Grafman%
}{%
Barbey%
\ \protect \BOthers {.}}{%
{\protect \APACyear {2013}}%
}]{%
barbey2013architecture}
\APACinsertmetastar {%
barbey2013architecture}%
\begin{APACrefauthors}%
Barbey, A\BPBI K.%
, Colom, R.%
\BCBL {}\ \BBA {} Grafman, J.%
\end{APACrefauthors}%
\unskip\
\newblock
\APACrefYearMonthDay{2013}{}{}.
\newblock
{\BBOQ}\APACrefatitle {Architecture of cognitive flexibility revealed by lesion mapping} {Architecture of cognitive flexibility revealed by lesion mapping}.{\BBCQ}
\newblock
\APACjournalVolNumPages{Neuroimage}{82}{}{547--554}.
\PrintBackRefs{\CurrentBib}

\bibitem [\protect \citeauthoryear {%
Baron-Cohen%
, Leslie%
\BCBL {}\ \BBA {} Frith%
}{%
Baron-Cohen%
\ \protect \BOthers {.}}{%
{\protect \APACyear {1985}}%
}]{%
baron1985does}
\APACinsertmetastar {%
baron1985does}%
\begin{APACrefauthors}%
Baron-Cohen, S.%
, Leslie, A\BPBI M.%
\BCBL {}\ \BBA {} Frith, U.%
\end{APACrefauthors}%
\unskip\
\newblock
\APACrefYearMonthDay{1985}{}{}.
\newblock
{\BBOQ}\APACrefatitle {Does the autistic child have a “theory of mind”?} {Does the autistic child have a “theory of mind”?}{\BBCQ}
\newblock
\APACjournalVolNumPages{Cognition}{21}{1}{37--46}.
\PrintBackRefs{\CurrentBib}

\bibitem [\protect \citeauthoryear {%
Beck%
, Robinson%
, Carroll%
\BCBL {}\ \BBA {} Apperly%
}{%
Beck%
\ \protect \BOthers {.}}{%
{\protect \APACyear {2006}}%
}]{%
beck2006children}
\APACinsertmetastar {%
beck2006children}%
\begin{APACrefauthors}%
Beck, S\BPBI R.%
, Robinson, E\BPBI J.%
, Carroll, D\BPBI J.%
\BCBL {}\ \BBA {} Apperly, I\BPBI A.%
\end{APACrefauthors}%
\unskip\
\newblock
\APACrefYearMonthDay{2006}{}{}.
\newblock
{\BBOQ}\APACrefatitle {Children's thinking about counterfactuals and future hypotheticals as possibilities} {Children's thinking about counterfactuals and future hypotheticals as possibilities}.{\BBCQ}
\newblock
\APACjournalVolNumPages{Child Development}{77}{2}{413--426}.
\PrintBackRefs{\CurrentBib}

\bibitem [\protect \citeauthoryear {%
Binz%
\ \BBA {} Schulz%
}{%
Binz%
\ \BBA {} Schulz%
}{%
{\protect \APACyear {2023}}%
}]{%
binz2023using}
\APACinsertmetastar {%
binz2023using}%
\begin{APACrefauthors}%
Binz, M.%
\BCBT {}\ \BBA {} Schulz, E.%
\end{APACrefauthors}%
\unskip\
\newblock
\APACrefYearMonthDay{2023}{}{}.
\newblock
{\BBOQ}\APACrefatitle {Using cognitive psychology to understand GPT-3} {Using cognitive psychology to understand gpt-3}.{\BBCQ}
\newblock
\APACjournalVolNumPages{Proceedings of the National Academy of Sciences (PNAS)}{120}{6}{e2218523120}.
\PrintBackRefs{\CurrentBib}

\bibitem [\protect \citeauthoryear {%
Borji%
}{%
Borji%
}{%
{\protect \APACyear {2023}}%
}]{%
borji2023categorical}
\APACinsertmetastar {%
borji2023categorical}%
\begin{APACrefauthors}%
Borji, A.%
\end{APACrefauthors}%
\unskip\
\newblock
\APACrefYearMonthDay{2023}{}{}.
\newblock
\APACrefbtitle {A categorical archive of chatgpt failures.} {A categorical archive of chatgpt failures.}
\PrintBackRefs{\CurrentBib}

\bibitem [\protect \citeauthoryear {%
Borke%
}{%
Borke%
}{%
{\protect \APACyear {1975}}%
}]{%
borke1975piaget}
\APACinsertmetastar {%
borke1975piaget}%
\begin{APACrefauthors}%
Borke, H.%
\end{APACrefauthors}%
\unskip\
\newblock
\APACrefYearMonthDay{1975}{}{}.
\newblock
{\BBOQ}\APACrefatitle {Piaget's mountains revisited: Changes in the egocentric landscape.} {Piaget's mountains revisited: Changes in the egocentric landscape.}{\BBCQ}
\newblock
\APACjournalVolNumPages{Developmental Psychology}{11}{2}{240}.
\PrintBackRefs{\CurrentBib}

\bibitem [\protect \citeauthoryear {%
Bubeck%
\ \protect \BOthers {.}}{%
Bubeck%
\ \protect \BOthers {.}}{%
{\protect \APACyear {2023}}%
}]{%
bubeck2023sparks}
\APACinsertmetastar {%
bubeck2023sparks}%
\begin{APACrefauthors}%
Bubeck, S.%
, Chandrasekaran, V.%
, Eldan, R.%
, Gehrke, J.%
, Horvitz, E.%
, Kamar, E.%
\BDBL {}Lundberg, S.%
\end{APACrefauthors}%
\unskip\
\newblock
\APACrefYearMonthDay{2023}{}{}.
\newblock
{\BBOQ}\APACrefatitle {Sparks of artificial general intelligence: Early experiments with gpt-4} {Sparks of artificial general intelligence: Early experiments with gpt-4}.{\BBCQ}
\newblock
\APACjournalVolNumPages{arXiv preprint arXiv:2303.12712}{}{}{}.
\PrintBackRefs{\CurrentBib}

\bibitem [\protect \citeauthoryear {%
Byrne%
}{%
Byrne%
}{%
{\protect \APACyear {2016}}%
}]{%
byrne2016counterfactual}
\APACinsertmetastar {%
byrne2016counterfactual}%
\begin{APACrefauthors}%
Byrne, R\BPBI M.%
\end{APACrefauthors}%
\unskip\
\newblock
\APACrefYearMonthDay{2016}{}{}.
\newblock
{\BBOQ}\APACrefatitle {Counterfactual thought} {Counterfactual thought}.{\BBCQ}
\newblock
\APACjournalVolNumPages{Annual Review of Psychology}{67}{}{135--157}.
\PrintBackRefs{\CurrentBib}

\bibitem [\protect \citeauthoryear {%
Byrne%
}{%
Byrne%
}{%
{\protect \APACyear {2017}}%
}]{%
byrne2017counterfactual}
\APACinsertmetastar {%
byrne2017counterfactual}%
\begin{APACrefauthors}%
Byrne, R\BPBI M.%
\end{APACrefauthors}%
\unskip\
\newblock
\APACrefYearMonthDay{2017}{}{}.
\newblock
{\BBOQ}\APACrefatitle {Counterfactual thinking: From logic to morality} {Counterfactual thinking: From logic to morality}.{\BBCQ}
\newblock
\APACjournalVolNumPages{Current Directions in Psychological Science}{26}{4}{314--322}.
\PrintBackRefs{\CurrentBib}

\bibitem [\protect \citeauthoryear {%
Chandra%
, Li%
, Tenenbaum%
\BCBL {}\ \BBA {} Ragan-Kelley%
}{%
Chandra%
\ \protect \BOthers {.}}{%
{\protect \APACyear {2023}}%
}]{%
chandra2023acting}
\APACinsertmetastar {%
chandra2023acting}%
\begin{APACrefauthors}%
Chandra, K.%
, Li, T\BHBI M.%
, Tenenbaum, J.%
\BCBL {}\ \BBA {} Ragan-Kelley, J.%
\end{APACrefauthors}%
\unskip\
\newblock
\APACrefYearMonthDay{2023}{}{}.
\newblock
{\BBOQ}\APACrefatitle {Acting as inverse inverse planning} {Acting as inverse inverse planning}.{\BBCQ}
\newblock
\BIn{} \APACrefbtitle {ACM SIGGRAPH Conference Proceedings.} {Acm siggraph conference proceedings.}
\PrintBackRefs{\CurrentBib}

\bibitem [\protect \citeauthoryear {%
Collins%
, Wong%
, Feng%
, Wei%
\BCBL {}\ \BBA {} Tenenbaum%
}{%
Collins%
\ \protect \BOthers {.}}{%
{\protect \APACyear {2022}}%
}]{%
collins2022structured}
\APACinsertmetastar {%
collins2022structured}%
\begin{APACrefauthors}%
Collins, K\BPBI M.%
, Wong, C.%
, Feng, J.%
, Wei, M.%
\BCBL {}\ \BBA {} Tenenbaum, J.%
\end{APACrefauthors}%
\unskip\
\newblock
\APACrefYearMonthDay{2022}{}{}.
\newblock
{\BBOQ}\APACrefatitle {Structured, flexible, and robust: benchmarking and improving large language models towards more human-like behavior in out-of-distribution reasoning tasks} {Structured, flexible, and robust: benchmarking and improving large language models towards more human-like behavior in out-of-distribution reasoning tasks}.{\BBCQ}
\newblock
\BIn{} \APACrefbtitle {Annual Meeting of the Cognitive Science Society (CogSci).} {Annual meeting of the cognitive science society (cogsci).}
\PrintBackRefs{\CurrentBib}

\bibitem [\protect \citeauthoryear {%
Dajani%
\ \BBA {} Uddin%
}{%
Dajani%
\ \BBA {} Uddin%
}{%
{\protect \APACyear {2015}}%
}]{%
dajani2015demystifying}
\APACinsertmetastar {%
dajani2015demystifying}%
\begin{APACrefauthors}%
Dajani, D\BPBI R.%
\BCBT {}\ \BBA {} Uddin, L\BPBI Q.%
\end{APACrefauthors}%
\unskip\
\newblock
\APACrefYearMonthDay{2015}{}{}.
\newblock
{\BBOQ}\APACrefatitle {Demystifying cognitive flexibility: Implications for clinical and developmental neuroscience} {Demystifying cognitive flexibility: Implications for clinical and developmental neuroscience}.{\BBCQ}
\newblock
\APACjournalVolNumPages{Trends in Neurosciences}{38}{9}{571--578}.
\PrintBackRefs{\CurrentBib}

\bibitem [\protect \citeauthoryear {%
De~Weerd%
, Verbrugge%
\BCBL {}\ \BBA {} Verheij%
}{%
De~Weerd%
\ \protect \BOthers {.}}{%
{\protect \APACyear {2017}}%
}]{%
de2017negotiating}
\APACinsertmetastar {%
de2017negotiating}%
\begin{APACrefauthors}%
De~Weerd, H.%
, Verbrugge, R.%
\BCBL {}\ \BBA {} Verheij, B.%
\end{APACrefauthors}%
\unskip\
\newblock
\APACrefYearMonthDay{2017}{}{}.
\newblock
{\BBOQ}\APACrefatitle {Negotiating with other minds: the role of recursive theory of mind in negotiation with incomplete information} {Negotiating with other minds: the role of recursive theory of mind in negotiation with incomplete information}.{\BBCQ}
\newblock
\APACjournalVolNumPages{Autonomous Agents and Multi-Agent Systems}{31}{}{250--287}.
\PrintBackRefs{\CurrentBib}

\bibitem [\protect \citeauthoryear {%
De~Weerd%
, Verbrugge%
\BCBL {}\ \BBA {} Verheij%
}{%
De~Weerd%
\ \protect \BOthers {.}}{%
{\protect \APACyear {2022}}%
}]{%
de2022higher}
\APACinsertmetastar {%
de2022higher}%
\begin{APACrefauthors}%
De~Weerd, H.%
, Verbrugge, R.%
\BCBL {}\ \BBA {} Verheij, B.%
\end{APACrefauthors}%
\unskip\
\newblock
\APACrefYearMonthDay{2022}{}{}.
\newblock
{\BBOQ}\APACrefatitle {Higher-order theory of mind is especially useful in unpredictable negotiations} {Higher-order theory of mind is especially useful in unpredictable negotiations}.{\BBCQ}
\newblock
\APACjournalVolNumPages{Autonomous Agents and Multi-Agent Systems}{36}{2}{30}.
\PrintBackRefs{\CurrentBib}

\bibitem [\protect \citeauthoryear {%
Epstude%
\ \BBA {} Roese%
}{%
Epstude%
\ \BBA {} Roese%
}{%
{\protect \APACyear {2008}}%
}]{%
epstude2008functional}
\APACinsertmetastar {%
epstude2008functional}%
\begin{APACrefauthors}%
Epstude, K.%
\BCBT {}\ \BBA {} Roese, N\BPBI J.%
\end{APACrefauthors}%
\unskip\
\newblock
\APACrefYearMonthDay{2008}{}{}.
\newblock
{\BBOQ}\APACrefatitle {The functional theory of counterfactual thinking} {The functional theory of counterfactual thinking}.{\BBCQ}
\newblock
\APACjournalVolNumPages{Personality and Social Psychology Review}{12}{2}{168--192}.
\PrintBackRefs{\CurrentBib}

\bibitem [\protect \citeauthoryear {%
Fan%
\ \protect \BOthers {.}}{%
Fan%
\ \protect \BOthers {.}}{%
{\protect \APACyear {2021}}%
}]{%
fan2021learning}
\APACinsertmetastar {%
fan2021learning}%
\begin{APACrefauthors}%
Fan, L.%
, Qiu, S.%
, Zheng, Z.%
, Gao, T.%
, Zhu, S\BHBI C.%
\BCBL {}\ \BBA {} Zhu, Y.%
\end{APACrefauthors}%
\unskip\
\newblock
\APACrefYearMonthDay{2021}{}{}.
\newblock
{\BBOQ}\APACrefatitle {Learning triadic belief dynamics in nonverbal communication from videos} {Learning triadic belief dynamics in nonverbal communication from videos}.{\BBCQ}
\newblock
\BIn{} \APACrefbtitle {Conference on Computer Vision and Pattern Recognition (CVPR).} {Conference on computer vision and pattern recognition (cvpr).}
\PrintBackRefs{\CurrentBib}

\bibitem [\protect \citeauthoryear {%
Fan%
, Xu%
, Cao%
, Zhu%
\BCBL {}\ \BBA {} Zhu%
}{%
Fan%
\ \protect \BOthers {.}}{%
{\protect \APACyear {2022}}%
}]{%
fan2022artificial}
\APACinsertmetastar {%
fan2022artificial}%
\begin{APACrefauthors}%
Fan, L.%
, Xu, M.%
, Cao, Z.%
, Zhu, Y.%
\BCBL {}\ \BBA {} Zhu, S\BHBI C.%
\end{APACrefauthors}%
\unskip\
\newblock
\APACrefYearMonthDay{2022}{}{}.
\newblock
{\BBOQ}\APACrefatitle {Artificial social intelligence: A comparative and holistic view} {Artificial social intelligence: A comparative and holistic view}.{\BBCQ}
\newblock
\APACjournalVolNumPages{CAAI Artificial Intelligence Research}{1}{2}{144--160}.
\PrintBackRefs{\CurrentBib}

\bibitem [\protect \citeauthoryear {%
Gergely%
, N{\'a}dasdy%
, Csibra%
\BCBL {}\ \BBA {} B{\'\i}r{\'o}%
}{%
Gergely%
\ \protect \BOthers {.}}{%
{\protect \APACyear {1995}}%
}]{%
gergely1995taking}
\APACinsertmetastar {%
gergely1995taking}%
\begin{APACrefauthors}%
Gergely, G.%
, N{\'a}dasdy, Z.%
, Csibra, G.%
\BCBL {}\ \BBA {} B{\'\i}r{\'o}, S.%
\end{APACrefauthors}%
\unskip\
\newblock
\APACrefYearMonthDay{1995}{}{}.
\newblock
{\BBOQ}\APACrefatitle {Taking the intentional stance at 12 months of age} {Taking the intentional stance at 12 months of age}.{\BBCQ}
\newblock
\APACjournalVolNumPages{Cognition}{56}{2}{165--193}.
\PrintBackRefs{\CurrentBib}

\bibitem [\protect \citeauthoryear {%
Herrmann%
, Call%
, Hern{\'a}ndez-Lloreda%
, Hare%
\BCBL {}\ \BBA {} Tomasello%
}{%
Herrmann%
\ \protect \BOthers {.}}{%
{\protect \APACyear {2007}}%
}]{%
herrmann2007humans}
\APACinsertmetastar {%
herrmann2007humans}%
\begin{APACrefauthors}%
Herrmann, E.%
, Call, J.%
, Hern{\'a}ndez-Lloreda, M\BPBI V.%
, Hare, B.%
\BCBL {}\ \BBA {} Tomasello, M.%
\end{APACrefauthors}%
\unskip\
\newblock
\APACrefYearMonthDay{2007}{}{}.
\newblock
{\BBOQ}\APACrefatitle {Humans have evolved specialized skills of social cognition: The cultural intelligence hypothesis} {Humans have evolved specialized skills of social cognition: The cultural intelligence hypothesis}.{\BBCQ}
\newblock
\APACjournalVolNumPages{Science}{317}{5843}{1360--1366}.
\PrintBackRefs{\CurrentBib}

\bibitem [\protect \citeauthoryear {%
Ionescu%
}{%
Ionescu%
}{%
{\protect \APACyear {2012}}%
}]{%
ionescu2012exploring}
\APACinsertmetastar {%
ionescu2012exploring}%
\begin{APACrefauthors}%
Ionescu, T.%
\end{APACrefauthors}%
\unskip\
\newblock
\APACrefYearMonthDay{2012}{}{}.
\newblock
{\BBOQ}\APACrefatitle {Exploring the nature of cognitive flexibility} {Exploring the nature of cognitive flexibility}.{\BBCQ}
\newblock
\APACjournalVolNumPages{New Ideas in Psychology}{30}{2}{190--200}.
\PrintBackRefs{\CurrentBib}

\bibitem [\protect \citeauthoryear {%
Kingsbury%
\ \BBA {} Hong%
}{%
Kingsbury%
\ \BBA {} Hong%
}{%
{\protect \APACyear {2020}}%
}]{%
kingsbury2020multi}
\APACinsertmetastar {%
kingsbury2020multi}%
\begin{APACrefauthors}%
Kingsbury, L.%
\BCBT {}\ \BBA {} Hong, W.%
\end{APACrefauthors}%
\unskip\
\newblock
\APACrefYearMonthDay{2020}{}{}.
\newblock
{\BBOQ}\APACrefatitle {A multi-brain framework for social interaction} {A multi-brain framework for social interaction}.{\BBCQ}
\newblock
\APACjournalVolNumPages{Trends in Neurosciences}{43}{9}{651--666}.
\PrintBackRefs{\CurrentBib}

\bibitem [\protect \citeauthoryear {%
Kosinski%
}{%
Kosinski%
}{%
{\protect \APACyear {2023}}%
}]{%
kosinski2023theory}
\APACinsertmetastar {%
kosinski2023theory}%
\begin{APACrefauthors}%
Kosinski, M.%
\end{APACrefauthors}%
\unskip\
\newblock
\APACrefYearMonthDay{2023}{}{}.
\newblock
{\BBOQ}\APACrefatitle {Theory of mind may have spontaneously emerged in large language models} {Theory of mind may have spontaneously emerged in large language models}.{\BBCQ}
\newblock
\APACjournalVolNumPages{arXiv preprint arXiv:2302.02083}{}{}{}.
\PrintBackRefs{\CurrentBib}

\bibitem [\protect \citeauthoryear {%
LeMare%
\ \BBA {} Rubin%
}{%
LeMare%
\ \BBA {} Rubin%
}{%
{\protect \APACyear {1987}}%
}]{%
lemare1987perspective}
\APACinsertmetastar {%
lemare1987perspective}%
\begin{APACrefauthors}%
LeMare, L\BPBI J.%
\BCBT {}\ \BBA {} Rubin, K\BPBI H.%
\end{APACrefauthors}%
\unskip\
\newblock
\APACrefYearMonthDay{1987}{}{}.
\newblock
{\BBOQ}\APACrefatitle {Perspective taking and peer interaction: Structural and developmental analyses} {Perspective taking and peer interaction: Structural and developmental analyses}.{\BBCQ}
\newblock
\APACjournalVolNumPages{Child Development}{}{}{306--315}.
\PrintBackRefs{\CurrentBib}

\bibitem [\protect \citeauthoryear {%
Liu%
, Fan%
, Rossi%
, Yao%
\BCBL {}\ \BBA {} Chen%
}{%
Liu%
\ \protect \BOthers {.}}{%
{\protect \APACyear {2016}}%
}]{%
liu2016effect}
\APACinsertmetastar {%
liu2016effect}%
\begin{APACrefauthors}%
Liu, H.%
, Fan, N.%
, Rossi, S.%
, Yao, P.%
\BCBL {}\ \BBA {} Chen, B.%
\end{APACrefauthors}%
\unskip\
\newblock
\APACrefYearMonthDay{2016}{}{}.
\newblock
{\BBOQ}\APACrefatitle {The effect of cognitive flexibility on task switching and language switching} {The effect of cognitive flexibility on task switching and language switching}.{\BBCQ}
\newblock
\APACjournalVolNumPages{International Journal of Bilingualism}{20}{5}{563--579}.
\PrintBackRefs{\CurrentBib}

\bibitem [\protect \citeauthoryear {%
X.~Ma%
, Gao%
\BCBL {}\ \BBA {} Xu%
}{%
X.~Ma%
\ \protect \BOthers {.}}{%
{\protect \APACyear {2023}}%
}]{%
ma2023tomchallenges}
\APACinsertmetastar {%
ma2023tomchallenges}%
\begin{APACrefauthors}%
Ma, X.%
, Gao, L.%
\BCBL {}\ \BBA {} Xu, Q.%
\end{APACrefauthors}%
\unskip\
\newblock
\APACrefYearMonthDay{2023}{}{}.
\newblock
{\BBOQ}\APACrefatitle {ToMChallenges: A Principle-Guided Dataset and Diverse Evaluation Tasks for Exploring Theory of Mind} {Tomchallenges: A principle-guided dataset and diverse evaluation tasks for exploring theory of mind}.{\BBCQ}
\newblock
\BIn{} \APACrefbtitle {Annual Meeting of the Association for Computational Linguistics (ACL).} {Annual meeting of the association for computational linguistics (acl).}
\PrintBackRefs{\CurrentBib}

\bibitem [\protect \citeauthoryear {%
Y.~Ma%
, Zhang%
\BCBL {}\ \BBA {} Zhu%
}{%
Y.~Ma%
\ \protect \BOthers {.}}{%
{\protect \APACyear {2023}}%
}]{%
ma2023brain}
\APACinsertmetastar {%
ma2023brain}%
\begin{APACrefauthors}%
Ma, Y.%
, Zhang, C.%
\BCBL {}\ \BBA {} Zhu, S\BHBI C.%
\end{APACrefauthors}%
\unskip\
\newblock
\APACrefYearMonthDay{2023}{}{}.
\newblock
{\BBOQ}\APACrefatitle {Brain in a Vat: On Missing Pieces Towards Artificial General Intelligence in Large Language Models} {Brain in a vat: On missing pieces towards artificial general intelligence in large language models}.{\BBCQ}
\newblock
\APACjournalVolNumPages{arXiv preprint arXiv:2307.03762}{}{}{}.
\PrintBackRefs{\CurrentBib}

\bibitem [\protect \citeauthoryear {%
Mitchell%
\ \BBA {} Krakauer%
}{%
Mitchell%
\ \BBA {} Krakauer%
}{%
{\protect \APACyear {2023}}%
}]{%
mitchell2023debate}
\APACinsertmetastar {%
mitchell2023debate}%
\begin{APACrefauthors}%
Mitchell, M.%
\BCBT {}\ \BBA {} Krakauer, D\BPBI C.%
\end{APACrefauthors}%
\unskip\
\newblock
\APACrefYearMonthDay{2023}{}{}.
\newblock
{\BBOQ}\APACrefatitle {The debate over understanding in AI’s large language models} {The debate over understanding in ai’s large language models}.{\BBCQ}
\newblock
\APACjournalVolNumPages{Proceedings of the National Academy of Sciences (PNAS)}{120}{13}{e2215907120}.
\PrintBackRefs{\CurrentBib}

\bibitem [\protect \citeauthoryear {%
Nyhout%
\ \BBA {} Ganea%
}{%
Nyhout%
\ \BBA {} Ganea%
}{%
{\protect \APACyear {2019}}%
}]{%
nyhout2019mature}
\APACinsertmetastar {%
nyhout2019mature}%
\begin{APACrefauthors}%
Nyhout, A.%
\BCBT {}\ \BBA {} Ganea, P\BPBI A.%
\end{APACrefauthors}%
\unskip\
\newblock
\APACrefYearMonthDay{2019}{}{}.
\newblock
{\BBOQ}\APACrefatitle {Mature counterfactual reasoning in 4-and 5-year-olds} {Mature counterfactual reasoning in 4-and 5-year-olds}.{\BBCQ}
\newblock
\APACjournalVolNumPages{Cognition}{183}{}{57--66}.
\PrintBackRefs{\CurrentBib}

\bibitem [\protect \citeauthoryear {%
OpenAI%
}{%
OpenAI%
}{%
{\protect \APACyear {2023}}%
}]{%
openai2023gpt4}
\APACinsertmetastar {%
openai2023gpt4}%
\begin{APACrefauthors}%
OpenAI.%
\end{APACrefauthors}%
\unskip\
\newblock
\APACrefYearMonthDay{2023}{}{}.
\newblock
{\BBOQ}\APACrefatitle {GPT-4 Technical Report} {Gpt-4 technical report}.{\BBCQ}
\newblock
\APACjournalVolNumPages{arXiv preprint arXiv:2303.08774}{}{}{}.
\PrintBackRefs{\CurrentBib}

\bibitem [\protect \citeauthoryear {%
Peng%
\ \protect \BOthers {.}}{%
Peng%
\ \protect \BOthers {.}}{%
{\protect \APACyear {2023}}%
}]{%
peng2023tong}
\APACinsertmetastar {%
peng2023tong}%
\begin{APACrefauthors}%
Peng, Y.%
, Han, J.%
, Zhang, Z.%
, Fan, L.%
, Liu, T.%
, Qi, S.%
\BDBL {}Zhu, S\BHBI C.%
\end{APACrefauthors}%
\unskip\
\newblock
\APACrefYearMonthDay{2023}{}{}.
\newblock
{\BBOQ}\APACrefatitle {The Tong Test: Evaluating Artificial General Intelligence Through Dynamic Embodied Physical and Social Interactions} {The tong test: Evaluating artificial general intelligence through dynamic embodied physical and social interactions}.{\BBCQ}
\newblock
\APACjournalVolNumPages{Engineering}{}{}{}.
\PrintBackRefs{\CurrentBib}

\bibitem [\protect \citeauthoryear {%
Rafetseder%
, Cristi-Vargas%
\BCBL {}\ \BBA {} Perner%
}{%
Rafetseder%
\ \protect \BOthers {.}}{%
{\protect \APACyear {2010}}%
}]{%
rafetseder2010counterfactual}
\APACinsertmetastar {%
rafetseder2010counterfactual}%
\begin{APACrefauthors}%
Rafetseder, E.%
, Cristi-Vargas, R.%
\BCBL {}\ \BBA {} Perner, J.%
\end{APACrefauthors}%
\unskip\
\newblock
\APACrefYearMonthDay{2010}{}{}.
\newblock
{\BBOQ}\APACrefatitle {Counterfactual reasoning: Developing a sense of “nearest possible world”} {Counterfactual reasoning: Developing a sense of “nearest possible world”}.{\BBCQ}
\newblock
\APACjournalVolNumPages{Child Development}{81}{1}{376--389}.
\PrintBackRefs{\CurrentBib}

\bibitem [\protect \citeauthoryear {%
Raffel%
\ \protect \BOthers {.}}{%
Raffel%
\ \protect \BOthers {.}}{%
{\protect \APACyear {2020}}%
}]{%
raffel2020exploring}
\APACinsertmetastar {%
raffel2020exploring}%
\begin{APACrefauthors}%
Raffel, C.%
, Shazeer, N.%
, Roberts, A.%
, Lee, K.%
, Narang, S.%
, Matena, M.%
\BDBL {}Liu, P\BPBI J.%
\end{APACrefauthors}%
\unskip\
\newblock
\APACrefYearMonthDay{2020}{}{}.
\newblock
{\BBOQ}\APACrefatitle {Exploring the limits of transfer learning with a unified text-to-text transformer} {Exploring the limits of transfer learning with a unified text-to-text transformer}.{\BBCQ}
\newblock
\APACjournalVolNumPages{Journal of Machine Learning Research (JMLR)}{21}{1}{5485--5551}.
\PrintBackRefs{\CurrentBib}

\bibitem [\protect \citeauthoryear {%
Sap%
, LeBras%
, Fried%
\BCBL {}\ \BBA {} Choi%
}{%
Sap%
\ \protect \BOthers {.}}{%
{\protect \APACyear {2022}}%
}]{%
sap2022neural}
\APACinsertmetastar {%
sap2022neural}%
\begin{APACrefauthors}%
Sap, M.%
, LeBras, R.%
, Fried, D.%
\BCBL {}\ \BBA {} Choi, Y.%
\end{APACrefauthors}%
\unskip\
\newblock
\APACrefYearMonthDay{2022}{}{}.
\newblock
{\BBOQ}\APACrefatitle {Neural theory-of-mind? on the limits of social intelligence in large lms} {Neural theory-of-mind? on the limits of social intelligence in large lms}.{\BBCQ}
\newblock
\BIn{} \APACrefbtitle {Annual Meeting of the Association for Computational Linguistics (ACL).} {Annual meeting of the association for computational linguistics (acl).}
\PrintBackRefs{\CurrentBib}

\bibitem [\protect \citeauthoryear {%
Schilbach%
\ \protect \BOthers {.}}{%
Schilbach%
\ \protect \BOthers {.}}{%
{\protect \APACyear {2013}}%
}]{%
schilbach2013toward}
\APACinsertmetastar {%
schilbach2013toward}%
\begin{APACrefauthors}%
Schilbach, L.%
, Timmermans, B.%
, Reddy, V.%
, Costall, A.%
, Bente, G.%
, Schlicht, T.%
\BCBL {}\ \BBA {} Vogeley, K.%
\end{APACrefauthors}%
\unskip\
\newblock
\APACrefYearMonthDay{2013}{}{}.
\newblock
{\BBOQ}\APACrefatitle {Toward a second-person neuroscience} {Toward a second-person neuroscience}.{\BBCQ}
\newblock
\APACjournalVolNumPages{Behavioral and Brain Sciences}{36}{4}{393--414}.
\PrintBackRefs{\CurrentBib}

\bibitem [\protect \citeauthoryear {%
Shiffrin%
\ \BBA {} Mitchell%
}{%
Shiffrin%
\ \BBA {} Mitchell%
}{%
{\protect \APACyear {2023}}%
}]{%
shiffrin2023probing}
\APACinsertmetastar {%
shiffrin2023probing}%
\begin{APACrefauthors}%
Shiffrin, R.%
\BCBT {}\ \BBA {} Mitchell, M.%
\end{APACrefauthors}%
\unskip\
\newblock
\APACrefYearMonthDay{2023}{}{}.
\newblock
{\BBOQ}\APACrefatitle {Probing the psychology of AI models} {Probing the psychology of ai models}.{\BBCQ}
\newblock
\APACjournalVolNumPages{Proceedings of the National Academy of Sciences (PNAS)}{120}{10}{e2300963120}.
\PrintBackRefs{\CurrentBib}

\bibitem [\protect \citeauthoryear {%
Sodian%
, Schoeppner%
\BCBL {}\ \BBA {} Metz%
}{%
Sodian%
\ \protect \BOthers {.}}{%
{\protect \APACyear {2004}}%
}]{%
sodian2004infants}
\APACinsertmetastar {%
sodian2004infants}%
\begin{APACrefauthors}%
Sodian, B.%
, Schoeppner, B.%
\BCBL {}\ \BBA {} Metz, U.%
\end{APACrefauthors}%
\unskip\
\newblock
\APACrefYearMonthDay{2004}{}{}.
\newblock
{\BBOQ}\APACrefatitle {Do infants apply the principle of rational action to human agents?} {Do infants apply the principle of rational action to human agents?}{\BBCQ}
\newblock
\APACjournalVolNumPages{Infant Behavior and Development}{27}{1}{31--41}.
\PrintBackRefs{\CurrentBib}

\bibitem [\protect \citeauthoryear {%
Stone%
}{%
Stone%
}{%
{\protect \APACyear {2006}}%
}]{%
stone2006theory}
\APACinsertmetastar {%
stone2006theory}%
\begin{APACrefauthors}%
Stone, V\BPBI E.%
\end{APACrefauthors}%
\unskip\
\newblock
\APACrefYearMonthDay{2006}{}{}.
\newblock
{\BBOQ}\APACrefatitle {Theory of mind and the evolution of social intelligence} {Theory of mind and the evolution of social intelligence}.{\BBCQ}
\newblock
\BIn{} \APACrefbtitle {Social neuroscience: People thinking about thinking people} {Social neuroscience: People thinking about thinking people}\ (\BPGS\ 103--129).
\newblock
\APACaddressPublisher{}{MIT Press}.
\PrintBackRefs{\CurrentBib}

\bibitem [\protect \citeauthoryear {%
Ullman%
}{%
Ullman%
}{%
{\protect \APACyear {2023}}%
}]{%
ullman2023large}
\APACinsertmetastar {%
ullman2023large}%
\begin{APACrefauthors}%
Ullman, T.%
\end{APACrefauthors}%
\unskip\
\newblock
\APACrefYearMonthDay{2023}{}{}.
\newblock
{\BBOQ}\APACrefatitle {Large language models fail on trivial alterations to theory-of-mind tasks} {Large language models fail on trivial alterations to theory-of-mind tasks}.{\BBCQ}
\newblock
\BIn{} \APACrefbtitle {Annual Meeting of the Association for Computational Linguistics (ACL).} {Annual meeting of the association for computational linguistics (acl).}
\PrintBackRefs{\CurrentBib}

\bibitem [\protect \citeauthoryear {%
Underwood%
\ \BBA {} Moore%
}{%
Underwood%
\ \BBA {} Moore%
}{%
{\protect \APACyear {1982}}%
}]{%
underwood1982perspective}
\APACinsertmetastar {%
underwood1982perspective}%
\begin{APACrefauthors}%
Underwood, B.%
\BCBT {}\ \BBA {} Moore, B.%
\end{APACrefauthors}%
\unskip\
\newblock
\APACrefYearMonthDay{1982}{}{}.
\newblock
{\BBOQ}\APACrefatitle {Perspective-taking and altruism.} {Perspective-taking and altruism.}{\BBCQ}
\newblock
\APACjournalVolNumPages{Psychological Bulletin}{91}{1}{143}.
\PrintBackRefs{\CurrentBib}

\bibitem [\protect \citeauthoryear {%
Wang%
, Wang%
, Paranamana%
\BCBL {}\ \BBA {} Shafto%
}{%
Wang%
\ \protect \BOthers {.}}{%
{\protect \APACyear {2020}}%
}]{%
wang2020mathematical}
\APACinsertmetastar {%
wang2020mathematical}%
\begin{APACrefauthors}%
Wang, P.%
, Wang, J.%
, Paranamana, P.%
\BCBL {}\ \BBA {} Shafto, P.%
\end{APACrefauthors}%
\unskip\
\newblock
\APACrefYearMonthDay{2020}{}{}.
\newblock
{\BBOQ}\APACrefatitle {A mathematical theory of cooperative communication} {A mathematical theory of cooperative communication}.{\BBCQ}
\newblock
\BIn{} \APACrefbtitle {Advances in Neural Information Processing Systems (NeurIPS).} {Advances in neural information processing systems (neurips).}
\PrintBackRefs{\CurrentBib}

\bibitem [\protect \citeauthoryear {%
Webb%
, Holyoak%
\BCBL {}\ \BBA {} Lu%
}{%
Webb%
\ \protect \BOthers {.}}{%
{\protect \APACyear {2023}}%
}]{%
webb2023emergent}
\APACinsertmetastar {%
webb2023emergent}%
\begin{APACrefauthors}%
Webb, T.%
, Holyoak, K\BPBI J.%
\BCBL {}\ \BBA {} Lu, H.%
\end{APACrefauthors}%
\unskip\
\newblock
\APACrefYearMonthDay{2023}{}{}.
\newblock
{\BBOQ}\APACrefatitle {Emergent analogical reasoning in large language models} {Emergent analogical reasoning in large language models}.{\BBCQ}
\newblock
\APACjournalVolNumPages{Nature Human Behaviour}{}{}{1--16}.
\PrintBackRefs{\CurrentBib}

\bibitem [\protect \citeauthoryear {%
Yakupov%
, Bakanova%
\BCBL {}\ \BBA {} Zhamieva%
}{%
Yakupov%
\ \protect \BOthers {.}}{%
{\protect \APACyear {2022}}%
}]{%
yakupov2022social}
\APACinsertmetastar {%
yakupov2022social}%
\begin{APACrefauthors}%
Yakupov, E\BPBI Z.%
, Bakanova, A\BPBI S.%
\BCBL {}\ \BBA {} Zhamieva, R\BPBI A.%
\end{APACrefauthors}%
\unskip\
\newblock
\APACrefYearMonthDay{2022}{}{}.
\newblock
{\BBOQ}\APACrefatitle {Social intelligence in the context of the development of subjective cognitive impairment} {Social intelligence in the context of the development of subjective cognitive impairment}.{\BBCQ}
\newblock
\APACjournalVolNumPages{Neurology Bulletin}{54}{3}{62--70}.
\PrintBackRefs{\CurrentBib}

\bibitem [\protect \citeauthoryear {%
Yang%
, Yu%
, Wang%
, Vong%
\BCBL {}\ \BBA {} Shafto%
}{%
Yang%
\ \protect \BOthers {.}}{%
{\protect \APACyear {2018}}%
}]{%
yang2018optimal}
\APACinsertmetastar {%
yang2018optimal}%
\begin{APACrefauthors}%
Yang, S\BPBI C\BHBI H.%
, Yu, Y.%
, Wang, P.%
, Vong, W\BPBI K.%
\BCBL {}\ \BBA {} Shafto, P.%
\end{APACrefauthors}%
\unskip\
\newblock
\APACrefYearMonthDay{2018}{}{}.
\newblock
{\BBOQ}\APACrefatitle {Optimal cooperative inference} {Optimal cooperative inference}.{\BBCQ}
\newblock
\BIn{} \APACrefbtitle {International Conference on Artificial Intelligence and Statistics (AISTATS).} {International conference on artificial intelligence and statistics (aistats).}
\PrintBackRefs{\CurrentBib}

\end{thebibliography}

\clearpage
\newpage
\newpage

\appendix

\twocolumn[
\begin{center}
    \LARGE \textbf{Appendix}
    \vspace{1cm}
\end{center}
]

\section{A. Dataset Generation}

\begin{figure}[!ht]
\centering    
    \subfloat[original scene]
    {
    \includegraphics[width=0.15\textwidth]{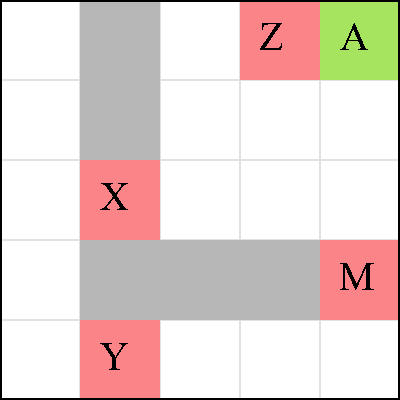}  
    }     
    \subfloat[\scalebox{0.7}{$Z>N>Y>M>X$}]
    { 
    \includegraphics[width=0.15\textwidth]{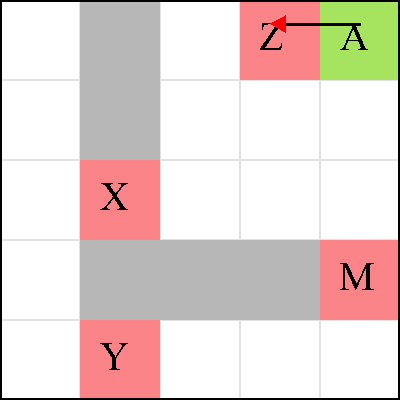}  
    }     
    \subfloat[\scalebox{0.7}{${M>X>Y>Z>N}$}]
    { 
    \includegraphics[width=0.15\textwidth]{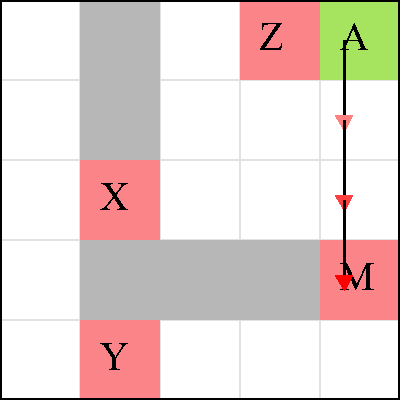}  
    }
    \\
    \subfloat[\scalebox{0.7}{$X>M>N>Z>Y$}]
    {
    \includegraphics[width=0.15\textwidth]{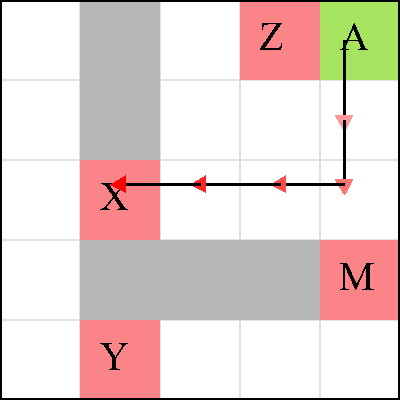}  
    }     
    \subfloat[\scalebox{0.7}{$Y>Z>N>M>X$}]
    { 
    \includegraphics[width=0.15\textwidth]{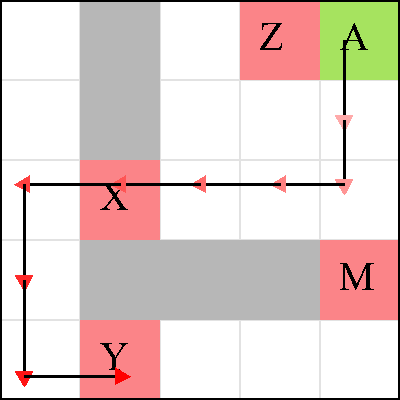}  
    }     
    \subfloat[\scalebox{0.7}{$N>Z>X>M>Y$}]
    { 
    \includegraphics[width=0.15\textwidth]{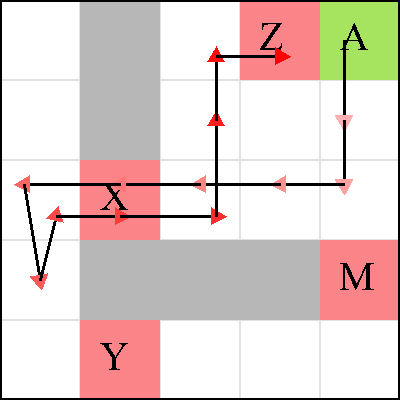}  
    } 
\caption{\textbf{Illustration of IR dataset generation.} For the same scenario configuration, different trajectories are obtained based on different preferences sampled randomly. We employ a greedy neighborhood search as our method for route planning.}
\label{fig:ft sample}
\end{figure}
\subsection{IR Dataset Generation} 

The generation of the entire IR dataset is detailed in \Cref{algo:ft_gen}. We randomly sample the initial scene (positions of obstacles, food truck slots and the agent), as well as the agent's rigid preference of the food trucks. Then, based on the initial state and the agent's rigid preference (as a strict total-order on food trucks), we construct the agent's whole trajectory. See an illustration of the process in \Cref{fig:ft sample}. Given the generated trajectory, we get the preference label via logical rules, which
is a strict partial-order preference describing all preferences that the trajectory could possibly indicate. As a special case, the original rigid preference satisfies the condition.
Theoretically, our synthesis algorithm can produce all possible cases.

\begin{algorithm}
\caption{IR dataset synthesis algorithm}
\label{algo:ft_gen}
\SetAlgoLined
\KwIn{$K$}
\KwOut{IR dataset of size $K$}
    \For{cnt $\gets 1$ \KwTo $K$}{
        \Do{not validCheck(\textbf{scene}) }{
            sample \textbf{scene} = (obstacles, agent, food trucks $(X,Y,Z,M)$)\;
        }
        sample \textbf{rigid preference} (e.g., $X > Z > M > N > Y$)\;
            \textbf{trj} $\gets$ planning(\textbf{scene}, \textbf{rigid preference})\; \tcp{neighbor search}
            \textbf{preference label} $\gets$ rule(\textbf{trj})\;
            \textbf{prompt} $\gets$ genPrompt(\textbf{scene}, \textbf{trj})\;
    }
\KwRet{scene, trj, rigid preference, preference label, prompt}
\end{algorithm}

\subsection{IIP Dataset Generation} 

The generation of the IIP dataset is summarized in \Cref{algo:iip_gen}. We also provide the detailed generation of each specific option: see \Cref{alg:option_misleading} for \nMisld{} option, \Cref{alg:option-far} for \nFar{} option, and \Cref{alg:option_best} for \nBest{} option. As for the \nShort{} option, we employ a trivial shortest route algorithm, and at the same time ensure it's distinct from other routes.

\begin{algorithm}
\caption{IIP dataset synthesis algorithm}
\label{algo:iip_gen} 
\SetAlgoLined
\KwIn{$K$}
\KwOut{IIP dataset of size $K$}
\For{cnt $\gets1$ \KwTo $K$}{
    \Do{not validCheck(\textbf{scene})}{
        sample \textbf{scene} = (obstacles, agent, restaurants $(X,Y)$)\;
    }
    \textbf{types}, \textbf{trjs} $\gets$ choicesGenerate(\textbf{scene})\; 
        \textbf{prompt} $\gets$ genPrompt(\textbf{scene}, \textbf{types}, \textbf{trjs})\;
}
\KwRet{scene, types, trjs, prompt}
\end{algorithm}

\begin{algorithm}
\caption{IIP route \nMisld{} generation}
\label{alg:option_misleading}
\SetAlgoLined
\KwIn{Scene S.}
\KwOut{Route \nMisld{} connecting $A$ to $X$.}
\textbf{Initialize} Color the scene $S$ based on \Cref{alg:coloring_iip}\; 
Construct a shortest route $\gamma_1$ from $A$ to $Y$\;
When multiple shortest routes exist for $\gamma_1$, choose the one avoiding $X$(red)-colored cells\;
Construct a shortest route $\gamma_2$ from $Y$ to $X$\;
\KwRet{the concatenation of $\gamma_1$ and $\gamma_2$}
\end{algorithm}

\begin{algorithm}
\caption{IIP route \nFar{} generation}
\label{alg:option-far}
\SetAlgoLined
\DontPrintSemicolon
\KwIn{Scene S.}
\KwOut{Route \nFar{} connecting $A$ to $X$.}
\BlankLine
\textbf{Initialize:} front queue $f=(Y,)$\;
\While{$f$ is nonempty}{
    Pop head $H$ of the queue\;
    Turn $H$ in the scene $S$ an obstacle\;
    \If{$A$ and $X$ are not connected in $S$ (i.e., there exists no route that connects $A$ and $X$)}{
        Turn $H$ accessible in $S$\;
    }
    Push adjacent roads of $H$ to $f$\;
}
\KwRet{the only route connecting $A$ to $X$}
\end{algorithm}

\begin{algorithm}
\caption{IIP route \nBest{} generation}
\label{alg:option_best}
\SetAlgoLined
\KwIn{Scene S.}
\KwOut{Route \nBest{} connecting $A$ to $X$.}
\textbf{Initialize} Color the scene $S$ based on \Cref{alg:coloring_iip}\; 
Find $X$(red)-colored cells $\Psi$ closest to $A$\;
\uIf {$|\Psi|=1$}{take $C$ as the unique element in $\Psi$\;}
\uElseIf{$|\Psi_1:=argmin_{\psi\in\Psi}(|\psi,X|)|=1$}{take $C$ as the unique element in $\Psi_1$\;}
\uElseIf{$|\Psi_2:=argmax_{\psi\in\Psi_1}(|\psi,Y|)|=1$}{take $C$ as the unique element in $\Psi_2$\;}
\Else{
{$\Psi_3:=argmax_{\psi\in\Psi_2}\cos(\overrightarrow{YX},\overrightarrow{A\psi})$\;}
{take $C$ as the unique element in $\Psi_3$ (uniqueness is quaranteed)\;}
}
Find a shortest route $\gamma_1$ from $A$ to $C$\;
Find a shortest route $\gamma_2$ from $C$ to $X$\;
When multiple shortest routes exist, choose the one far from $Y$(blue)-colored region\;
\KwRet{the concatenation of $\gamma_1$ and $\gamma_2$}
\end{algorithm}

\begin{figure}[H]
\centering    
    \subfloat[unstained]
    {
    \includegraphics[width=0.15\textwidth]{ 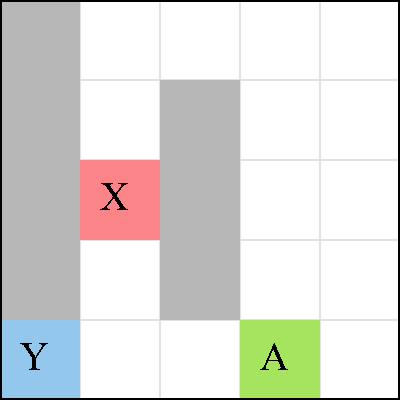}  
    }     
    \subfloat[full coloring]
    { 
    \includegraphics[width=0.15\textwidth]{ 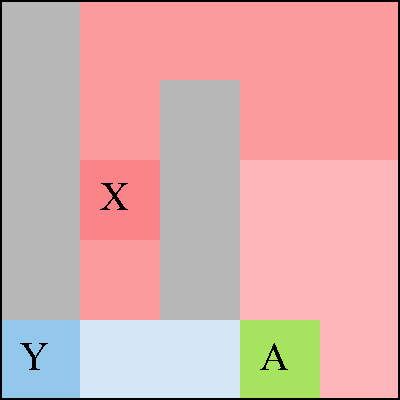}  
    }
    \subfloat[\nBest{} option]
    { 
    \includegraphics[width=0.15\textwidth]{ 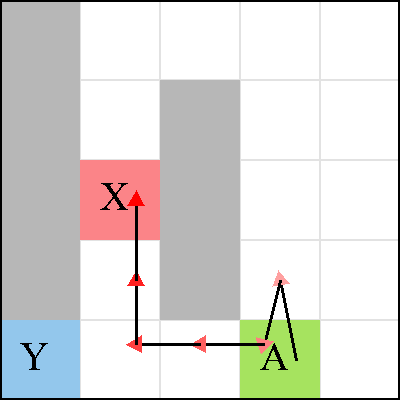}  
    }
\caption{\textbf{Illustration of IIP coloring strategy.} (a) Original unstained scene. (b) Fully colored scene. (c) \nBest{} route.}
\label{fig:iip coloring}
\end{figure}

The coloring algorithm is detailed in \Cref{alg:coloring_iip}. For colors  $(C, k)$, $C\in\{X,Y,N\}$ represents different possible preferences (reflected in different colors, i.e., red for $X$, blue for $Y$, and white for $N$), $N$ represents neutral preference, and $k\ge0$ represents the preference signaling strength via color intensity $e^{-\beta k}$ in the model. \Cref{fig:iip coloring} also provides an example of the coloring algorithm.

\begin{algorithm}
\caption{Coloring strategy in IIP}
\label{alg:coloring_iip}
\SetAlgoLined
\KwIn{Scene S, distance measure $|a,b|$ (length of shortest path), colors $(C, k)$, where $C\in\{X,Y,N\}$, $N$ represents neutral preference, $k\ge0$.}
\KwOut{Coloring of each cell in $S$.}
\textbf{Initialize} $r_X^0=r_Y^0=\emptyset$, $r_X^1=\!\{X\}$, $r_Y^1=\!\{Y\}$, $k\!=\!1$\; 
\While{$(r_X^k-r_X^{k-1})\cup (r_Y^k-r_Y^{k-1})\ne\emptyset$}{
    Color $r_X^k$ by $(X, k)$, color $r_Y^k$ by $(Y,k)$\;
    $r_X^{k+1}\!\gets\!\{Z\!:\!|Z,\!r_X^{k}|\!-\!|A,\!r_X^{k}|\!<\!|Z,\!r_Y^{k}|\!-\!|A,\! r_Y^{k}|\}$\footnotemark\;
    $r_Y^{k+1}\!\gets\!\{Z\!:\!|Z,\!r_Y^{k}|\!-\!|A,\!r_Y^{k}|\!<\!|Z,\!r_X^{k}|\!-\!|A,\! r_X^{k}|\}$ \;
    $k\gets k+1$\;
}
Color all uncolored cells by $(N,0)$.

\KwRet{Fully colored scene.}
\end{algorithm}

\footnotetext{Equivalent to $\!|A,Z\!|+\!|Z,\!r_X^{k}|\!-\!|A,\!r_X^{k}|\!<\!|A,Z\!|+\!|Z,\!r_Y^{k}|\!-\!|A,\! r_Y^{k}|$.}

\section{B. Prompt}

\subsection{IR Zero-shot Prompt}

\begin{figure}[h]
\centering    
    \subfloat[IR task]
    { 
    \includegraphics[width=0.22\textwidth]{ 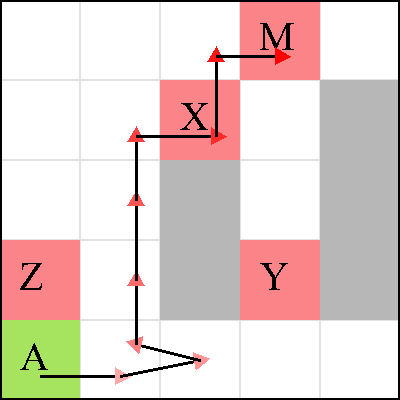}  
    }
    \subfloat[IIP scene]
    {
    \includegraphics[width=0.22\textwidth]{ 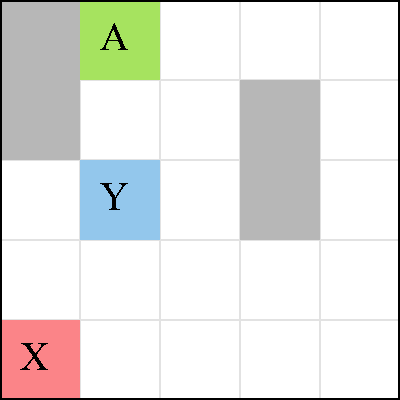}  
    }  
    \\ 
    \subfloat[IIP Route A]
    {
    \includegraphics[width=0.11\textwidth]{ 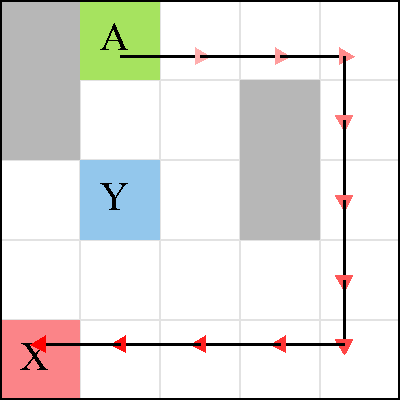}  
    }  
    \subfloat[IIP Route B]
    {
    \includegraphics[width=0.11\textwidth]{ 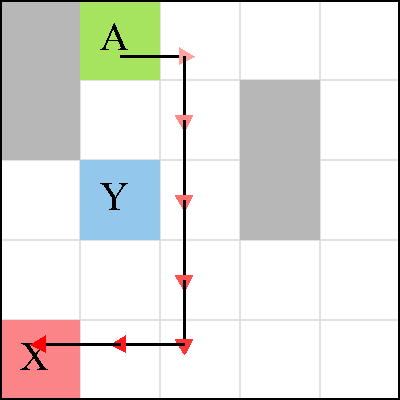}  
    }  
    \subfloat[IIP Route C]
    {
    \includegraphics[width=0.11\textwidth]{ 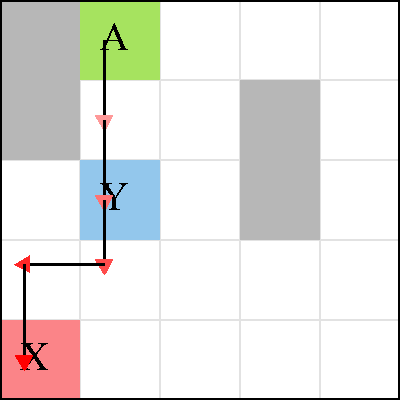}  
    }  
    \subfloat[IIP Route D]
    {
    \includegraphics[width=0.11\textwidth]{ 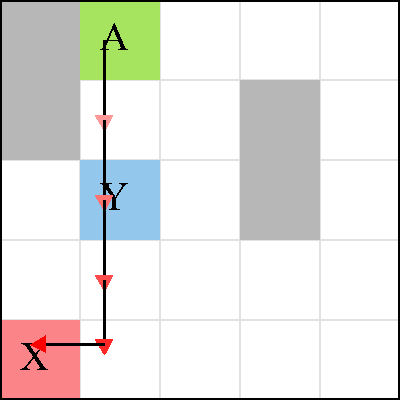}  
    }  
\caption{\textbf{Illustration stimuli examples of zero-shot cases for IR and IIP.} (a) Scene layout and actor’s trajectory in the IR task; (b) Scene layout for the IIP task; (c)-(f) Four potential routes for the actor in the IIP task scenario (b).}
\label{fig:zeroshots}
\end{figure}

We use the following prompt in zero-shot IR task, corresponding to the image depicted in \Cref{fig:zeroshots}(a).

\begin{prompt}
Question:
Please follow the instructions to answer the question. Below is one possible layout of the food truck area. The letter `A' stands for Student A, `*' stands for empty areas, and `W' stands for obstructed walls that block the student. Other letters represent different kinds of food.\\
\\
We're assuming the top left corner is (0,0), top right is (4,0), bottom left is (0,4), and bottom right is (4,4). Here is student A's trajectory. The coordinates reflect the position of the A. Each time student A can move one step.\\
\\
Layout:\\
***M*\\
**X*W\\
**W*W\\
Z*WYW\\
A****\\
\\
\\
Student A's Trajectory:\\
Here is the student A's trajectory. The coordinates reflect the position of the A. Each time agent can move one step.\\
(0, 4) view Z; memory Z\\
(1, 4) view Z; memory Z\\
(2, 4) view Y; memory Z,Y\\
(1, 4) view Z; memory Z,Y\\
(1, 3) view Z; memory Z,Y\\
(1, 2) view X,Z; memory Z,Y,X\\
(1, 1) view X; memory Z,Y,X\\
(2, 1) view X,M; memory Z,Y,X,M\\
(2, 0) view X,M; memory Z,Y,X,M\\
(3, 0) view X,M; memory Z,Y,X,M; pick M\\
\\
Please determine the preference among all the five foods foods and provide your answer following the format.
\end{prompt}

\subsection{IR Few-shot Prompt} 

\begin{figure}[H]
\centering    
    \subfloat[Trajectory 1]
    { 
    \includegraphics[width=0.15\textwidth]{ 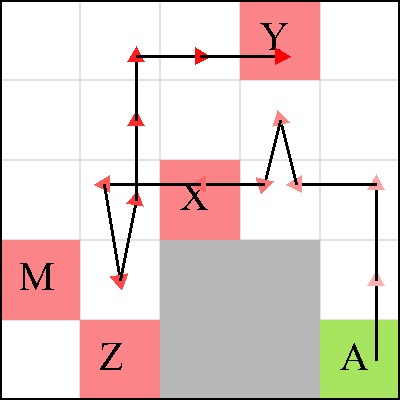}  
    }
    \subfloat[Trajectory 2]
    {
    \includegraphics[width=0.15\textwidth]{ 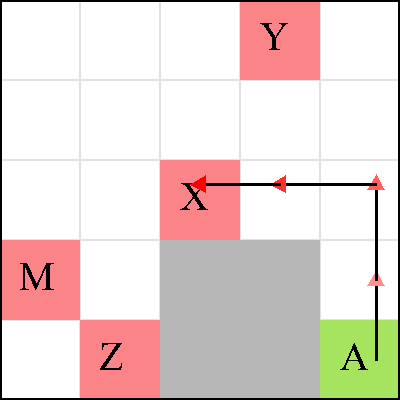}  
    }     
    \subfloat[Trajectory 3]
    { 
    \includegraphics[width=0.15\textwidth]{ 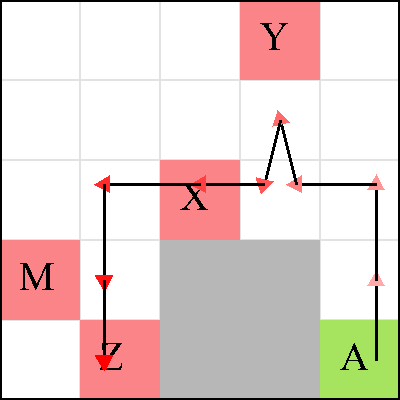}  
    }
\caption{\textbf{Illustration of three IR few-shot cases.} (a) \nTurn{} (b) \nDirect{} (c) \nFinal{}.}
\label{fig:FT oneshot}
\end{figure}

The following prompt demonstrates several few-shot cases for the IR task, with the visualization of their trajectories in \Cref{fig:FT oneshot}. For human subjects, we only provided Trajectory 1, which is the \nTurn{} case.

\begin{prompt}
You will be presented with three examples which share the same layout to solve the problem. Please go through the example carefully to understand the solution. Here is a layout and the trajectory of student A. We're assuming the top left corner is (0,0), top right is (4,0), bottom left is (0,4), and bottom right is (4,4). Here is student A's trajectory. The coordinates reflect the position of the A. Each time student A can move one step.\\
\\
Layout:\\
***Y*\\
*****\\
**X**\\
M*WW*\\
*ZWWA\\
\\
Student A's Trajectory 1:\\
Here is the student A's trajectory. The coordinates reflect the position of the A. Each time agent can move one step.\\
(4, 4)\\
(4, 3)\\
(4, 2)\\
(3, 2) view X; memory X\\
(3, 1) view X,Y; memory X,Y\\
(3, 2) view X; memory X,Y\\
(2, 2) view X; memory X,Y\\
(1, 2) view X,M; memory X,Y,M\\
(1, 3) view X,Z,M; memory X,Y,M,Z\\
(1, 2) view X,M; memory X,Y,M,Z\\
(1, 1) view X; memory X,Y,M,Z\\
(1, 0) memory X,Y,M,Z\\
(2, 0) view Y; memory X,Y,M,Z\\
(3, 0) view Y; memory X,Y,M,Z; pick Y\\
\\
Answer 1: \\
N>Y>\{X,Z,M\}\\
Explanation 1: \\
When Student A explores all the food options and then goes back to choose Y, it implies that Y is his second favorite food. This suggests that Student A's favorite food is not available today, as he would not have returned to pick up his second favorite otherwise.\\
\\
\\
Student A's Trajectory 2:\\
Here is the student A's trajectory. The coordinates reflect the position of the A. Each time agent can move one step.\\
(4, 4)\\
(4, 3)\\
(4, 2)\\
(3, 2) view X; memory X\\
(2, 2) view X; memory X; pick X\\
\\
Answer 2: \\
X > \{M,N,Y,Z\}\\
Explanation 2: \\
Student A picks up X without fully exploring other options, suggesting that X is his favorite food, while his preferences for other options remain unknown.\\
\\
\\
Student A's Trajectory 3:\\
Here is the student A's trajectory. The coordinates reflect the position of the A. Each time agent can move one step.\\
(4, 4)\\
(4, 3)\\
(4, 2)\\
(3, 2) view X; memory X\\
(3, 1) view X,Y; memory X,Y\\
(3, 2) view X; memory X,Y\\
(2, 2) view X; memory X,Y\\
(1, 2) view X,M; memory X,Y,M\\
(1, 3) view X,Z,M; memory X,Y,M,Z\\
(1, 4) view Z,M; memory X,Y,M,Z; pick Z\\
\\
Answer 3: \\
Z > \{M,X,Y\}, \{N\}\\
Explanation 3: \\
Student A thoroughly examines all the available options and ultimately selects option Z. This suggests that he prefers Z over the other alternatives—X, Y, and M. However, his preference for option N remains unclear. It is possible that Z is his favorite food, or alternatively, N could be his favorite food. In the latter case, due to N's unavailability, he might have opted for his second favorite choice, Y.
\end{prompt}

\subsection{IIP Zero-shot Prompt}

We use the following prompt in the zero-shot IIP task, corresponding to the images depicted in \Cref{fig:zeroshots}(b-f).

\begin{prompt}
Setting:\\
A campus area is represented by a 5*5 grid. There are only two restaurants, X and Y on the campus. Student A attends school daily and is fully aware of the locations of each restaurant. He has a clear pre-established preference between X and Y, that is, he decides to eat at restaurant X. Observer B is an observer who monitors A's actions and is smart enough to infer A's preference once it has been signaled.\\
\\
Action:\\
Student A can only take one step each time in four directions: up, down, left, and right. He wants to carefully plan his actions to achieve two goals.\\
Primary goal: He wants to signal his preference (Restaurant X) to B as early as possible with the least ambiguity.\\
Secondary goal: Once he thinks that the preference has been signaled, he will move to Restaurant X as soon as possible because he is hungry.\\
\\
Layout:\\
Below is one possible layout of the campus area. The letter `A' stands for Student A, `*' stands for empty areas, and `W' stands for obstructed walls that block the student. The top-left grid cell is designated as (0,0), the top-right as (4,0), the bottom-left as (0,4), and the bottom-right as (4,4). The letters `X' and `Y' stand for two restaurants.\\
WA***\\
W**W*\\
*Y*W*\\
*****\\
X****\\
\\
Task:\\
Your task is to help A to choose the optimal action trajectory to achieve the above goals. Also, calculate the number of steps required to achieve the primary goal.\\
\\
Question: Most Proper Route\\
Route A\\
Move right from (1, 0) to (2,0)\\
Move right from (2, 0) to (3,0)\\
Move right from (3, 0) to (4,0)\\
Move down from (4, 0) to (4,1)\\
Move down from (4,1) to (4,2)\\
Move down from (4, 2) to (4,3)\\
Move down from (4, 3) to (4,4)\\
Move left from (4, 4) to (3,4)\\
Move left from (3, 4) to (2,4)\\
Move left from (2, 4) to (1,4)\\
Move left from (1, 4) to (0,4)\\
\\
Route B\\
Move right from (1, 0) to (2,0)\\
Move down from (2, 0) to (2,1)\\
Move down from (2, 1) to (2,2)\\
Move down from (2, 2) to (2,3)\\
Move down from (2, 3) to (2,4)\\
Move left from (2, 4) to (1,4)\\
Move left from (1, 4) to (0,4)\\
\\
Route C\\
Move down from (1,0) to (1,1)\\
Move down from (1,1) to (1,2)\\
Move down from (1,2) to (1,3)\\
Move left from (1,3) to (0,3)\\
Move down from (0,3) to (0,4)\\
\\
Route D\\
Move down from (1,0) to (1,1)\\
Move down from (1,1) to (l,2)\\
Move down from (1,2) to (l,3)\\
Move down from (1,3) to (1,4)\\
Move left from (1,4) to (0,4)\\
\end{prompt}

\subsection{IIP Few-shot Prompt} 

\begin{figure}[H]
\centering    
    \subfloat[Scene]
    { 
    \includegraphics[width=0.15\textwidth]{ 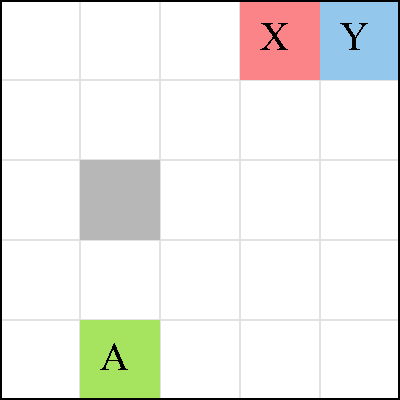}  
    }
    \subfloat[Route A(\nBest)]
    {
    \includegraphics[width=0.15\textwidth]{ 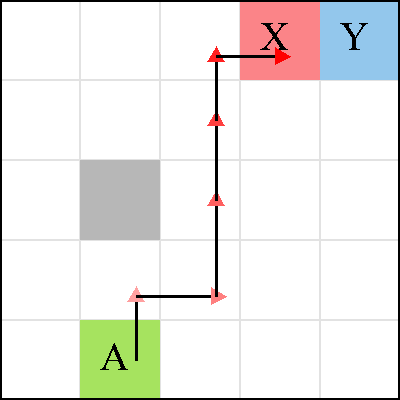}  
    } 
    \\
    \subfloat[Route B(\nFar)]
    { 
    \includegraphics[width=0.15\textwidth]{ 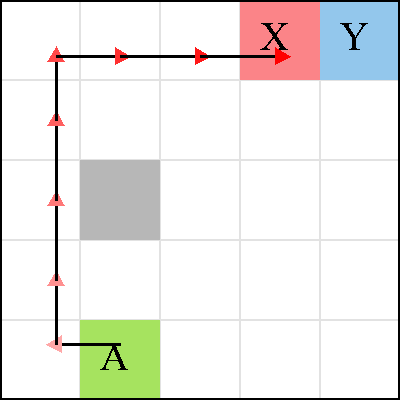}  
    }
    \subfloat[Route C(\nMisld)]
    {
    \includegraphics[width=0.15\textwidth]{ 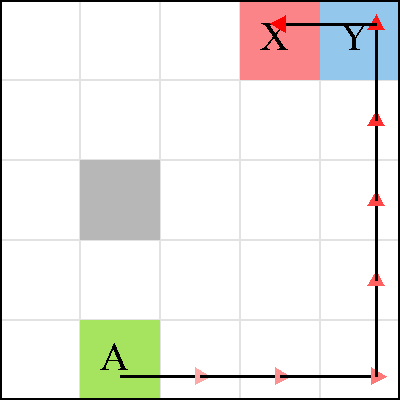}  
    }     
    \subfloat[Route D(\nShort)]
    { 
    \includegraphics[width=0.15\textwidth]{ 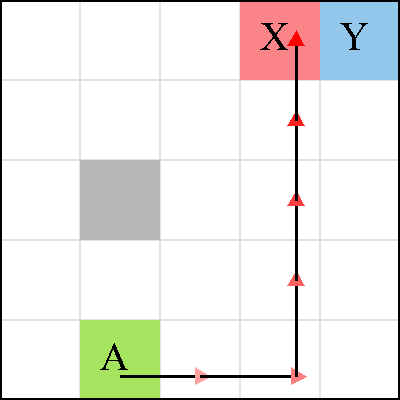}  
    }
\caption{\textbf{Illustration of IIP few-shot cases.}}
\label{fig:iip oneshot}
\end{figure}

The graphical version of the problem scenario and each option can be seen in \Cref{fig:iip oneshot}.

\begin{prompt}
Example: \\
Below is one possible setting of the campus area. Student A is at (1,4) and Restaurant X is at (3,0)
Route A:\\
Start at (1,4), go up to (1,3), then right to (2,3). Continue up to (2,0) and finally right to X (3,0). This route indicates a preference for X (3,0) by initially moving upwards. This avoids any suggestion of heading towards Y (4,0) that could be inferred from a rightward movement. Once the preference is signaled, the route then opts for the shortest route.\\
Route B:\\
Begin at (1,4), move left to (0,4), and go up to (0,0). Then move right to X (3,0). This route moves left first and continues to bypass the wall from the left to avoid the misinterpretation of intention during the whole movement. \\
Route C:\\
Start at (1,4), go right to (4,4), then up to Y (4,0) and left to X (3,0). This route only indicates that the target is X (3,0) not Y (4,0) when moving away from Y after it reaches Y.\\
Route D:\\
From (1,4), move right to (3,4), then up to X (3,0). This is a simple, direct route to X (3,0). \\
\\
As you may have realized, our routes in each problem are of the above 4 styles but occur in each problem in randomly shuffled orders.
\end{prompt}

\section{C. Evaluation Criteria}

\begin{table}[H]
\centering
\caption{Cognitive abilities reflected in IR and IIP. R: Rationality, C: Counterfactual reasoning, P: Perspective switching, F: Cognitive flexibility.}
\label{tab: cognitive abilities}
\begin{tabular}{lcccc}
\toprule
\small 
& \circled{R} & \circled{C} & \circled{P} & \circled{F} \\
\midrule
IR          &  \ding{51} & \ding{51} & \ding{51} & \ding{55} \\
IIP-\nShort &  \ding{51} & \ding{55} & \ding{55} & \ding{55} \\
IIP-\nMisld &  \ding{51} & \ding{51} & \ding{51} & \ding{55} \\
IIP-\nFar   &  \ding{51} & \ding{51} & \ding{51} & \ding{55} \\
IIP-\nBest  &  \ding{51} & \ding{51} & \ding{51} & \ding{51} \\
\bottomrule
\end{tabular}
\end{table}


\Cref{tab: cognitive abilities} provides a qualitative analysis of cognitive abilities reflected in our two tasks. 

\section{D. Human Study}

We carried out experiments involving human participants using the Qualtrics\footnote{\url{https://www.qualtrics.com/}} platform, with the respective online URLs as follows. The text-only version or the with-image version tests are randomly distributed.

\begin{itemize}
    \item IR survey: \url{https://bnupsych.asia.qualtrics.com/jfe/form/SV_baurQ9tSwFQayVM} 
    \item IIP survey: \url{https://bnupsych.asia.qualtrics.com/jfe/form/SV_6FoGehYJNCoIVlY}
\end{itemize}

\subsection{Statistical Hypothesis Testing}

\begin{table*}[h]
\centering
\caption{\textbf{Multiple Hypothesis Testing Results in Human Studies on the IR Task.}}
\label{tab:multiple_hypothesis_test_ft}
\begin{tabular}{lllllll}
\toprule
\textbf{Test} & \textbf{H0} & \textbf{H1} & \textbf{Method} & \textbf{Test Stats} & \textbf{P-Value} & \textbf{Conclusion} \\
\midrule
T2I on ZS & $\mu_1=\mu_2$ & $\mu_1\neq\mu_2$ &  T-test & t=0.7409 & 0.4612 & Fail to reject H0 \\
T2I on OS & $\mu_1=\mu_2$ & $\mu_1\neq\mu_2$ & T-test & t=-0.2840 & 0.7772 & Fail to reject H0 \\
T2I on ZS and \nDirect & $\mu_1=\mu_2$ & $\mu_1\neq\mu_2$ & T-test & t=1.4671 & 0.1467 & Fail to reject H0 \\
T2I on ZS and \nFinal & $\mu_1=\mu_2$ & $\mu_1\neq\mu_2$ & T-test & t=0.4507 & 0.6536 & Fail to reject H0 \\
T2I on ZS and \nTurn & $\mu_1=\mu_2$ & $\mu_1\neq\mu_2$ & T-test & t=-0.1441 & 0.8858 & Fail to reject H0 \\
T2I on OS and \nDirect & $\mu_1=\mu_2$ & $\mu_1\neq\mu_2$ & T-test & t=-0.3633 & 0.7182 & Fail to reject H0 \\
T2I on OS and \nFinal & $\mu_1=\mu_2$ & $\mu_1\neq\mu_2$ & T-test & t=0.5753 & 0.5679 & Fail to reject H0 \\
T2I on OS and \nTurn & $\mu_1=\mu_2$ & $\mu_1\neq\mu_2$ & T-test & t=-0.8751 & 0.3875 & Fail to reject H0 \\
Types on Text and ZS & $\mu_1=\mu_2$ & $\mu_1\neq\mu_2$ & One-way ANOVA & f=6.8661 & 0.0015 & Reject H0 \\
Types on Image and ZS & $\mu_1=\mu_2$ & $\mu_1\neq\mu_2$ & One-way ANOVA & f=5.5072 & 0.0053 & Reject H0 \\
Types on Text and OS & $\mu_1=\mu_2$ & $\mu_1\neq\mu_2$ & One-way ANOVA & f=4.2459 & 0.0184 & Reject H0 \\
Types on Image and OS & $\mu_1=\mu_2$ & $\mu_1\neq\mu_2$ & One-way ANOVA & f=9.9272 & 0.0002 & Reject H0 \\
Z2O on Text & $\mu_1=\mu_2$ & $\mu_1\neq\mu_2$ & T-test & t=-1.8633 & 0.0664 & Fail to reject H0 \\
Z2O on Text and \nDirect & $\mu_1=\mu_2$ & $\mu_1\neq\mu_2$ & T-test & t=-0.6162 & 0.5401 & Fail to reject H0 \\
Z2O on Text and \nFinal& $\mu_1=\mu_2$ & $\mu_1\neq\mu_2$ & T-test & t=-1.5737 & 0.1208 & Fail to reject H0 \\
Z2O on Text and \nTurn & $\mu_1=\mu_2$ & $\mu_1\neq\mu_2$ & T-test & t=-2.5089 & 0.0150 & Reject H0 \\
Z2O on Image & $\mu_1=\mu_2$ & $\mu_1\neq\mu_2$ & T-test & t=-2.7387 & 0.0078 & Reject H0 \\
Z2O on Image and \nDirect & $\mu_1=\mu_2$ & $\mu_1\neq\mu_2$ & T-test & t=-2.0984 & 0.0407 & Reject H0 \\
Z2O on Image and \nFinal & $\mu_1=\mu_2$ & $\mu_1\neq\mu_2$ & T-test & t=-1.2466 & 0.2176 & Fail to reject H0 \\
Z2O on Image and \nTurn & $\mu_1=\mu_2$ & $\mu_1\neq\mu_2$ & T-test & t=-3.1492 & 0.0027 & Reject H0 \\
Warmup on ZS and Text& $\mu_1=\mu_2$ & $\mu_1\neq\mu_2$ & T-test & t=-0.5853 & 0.5601 & Fail to reject H0 \\
Warmup on ZS and Image& $\mu_1=\mu_2$ & $\mu_1\neq\mu_2$ & T-test & t=-0.4107 & 0.6825 & Fail to reject H0 \\
\bottomrule
\end{tabular}
\end{table*}

\begin{table*}[h]
\centering
\caption{\textbf{Multiple Hypothesis Testing Results in Human Studies on the IIP Task.}}
\label{tab:multiple_hypothesis_test_iip}
\begin{tabular}{lllllll}
\toprule
\textbf{Test} & \textbf{H0} & \textbf{H1} & \textbf{Method} & \textbf{Test Stats} & \textbf{P-Value} & \textbf{Conclusion} \\
\midrule
T2I on ZS & equivalent & not equivalent  & Chi-square test & $\chi^2$=6.3549 & 0.0956 & Fail to reject H0 \\
T2I on OS & equivalent & not equivalent  & Chi-square test & $\chi^2$=0.7795 & 0.7795 & Fail to reject H0 \\
Z2O on Text & equivalent & not equivalent  & Chi-square test & $\chi^2$=2.8473 & 0.4158 & Fail to reject H0 \\
Z2O on Image & equivalent & not equivalent  & Chi-square test & $\chi^2$=0.1444 & 0.9860 & Fail to reject H0 \\
Types on Text and ZS & equivalent & not equivalent  & Chi-square test & $\chi^2$=54.0807 & 0.0000 & Reject H0 \\
Types on Image and ZS & equivalent & not equivalent  & Chi-square test & $\chi^2$=36.4588 & 0.0000 & Reject H0 \\
Types on Text and OS & equivalent & not equivalent  & Chi-square test & $\chi^2$=32.6774 & 0.0002 & Reject H0 \\
Types on Image and OS & equivalent & not equivalent  & Chi-square test & $\chi^2$=26.2358 & 0.0019 & Reject H0 \\

\bottomrule
\end{tabular}
\end{table*}

As shown in \Cref{tab:multiple_hypothesis_test_ft,tab:multiple_hypothesis_test_iip}, we conducted multiple hypothesis testing for each of the two tasks in human study. ``T2I'' means ``text vs. image'', ``ZS'' means ``zero shot'', ``OS'' means ``one shot'', ``Z2O'' means ``zero shot vs. one shot'', ``Warm-up'' means dividing the six questions before the zero-shot test into two groups based on chronological order, and comparing the statistical significance between the earlier and later groups. The \Cref{tab:multiple_hypothesis_test_ft} clearly indicates that significant differences among human subjects are only present between the types (\nDirect/\nFinal/\nFinal) of the IR task. They are not sensitive to text or images, whether they have undergone an one-shot, or to the order of answering the questions. The same conclusion applies to the IIP task, where the ``type'' in IIP refers to Type I-IV.

\section{E. Additional Experiments}

\subsection{Generalization Test in IR task}

As \ac{llm} usually reported 
to have an enhanced capability 
under in-context learning, we designed several in-context (few-shots) tests.

\begin{figure*}[t!]
\centering    
    { 
    \includegraphics[width=0.95\textwidth]{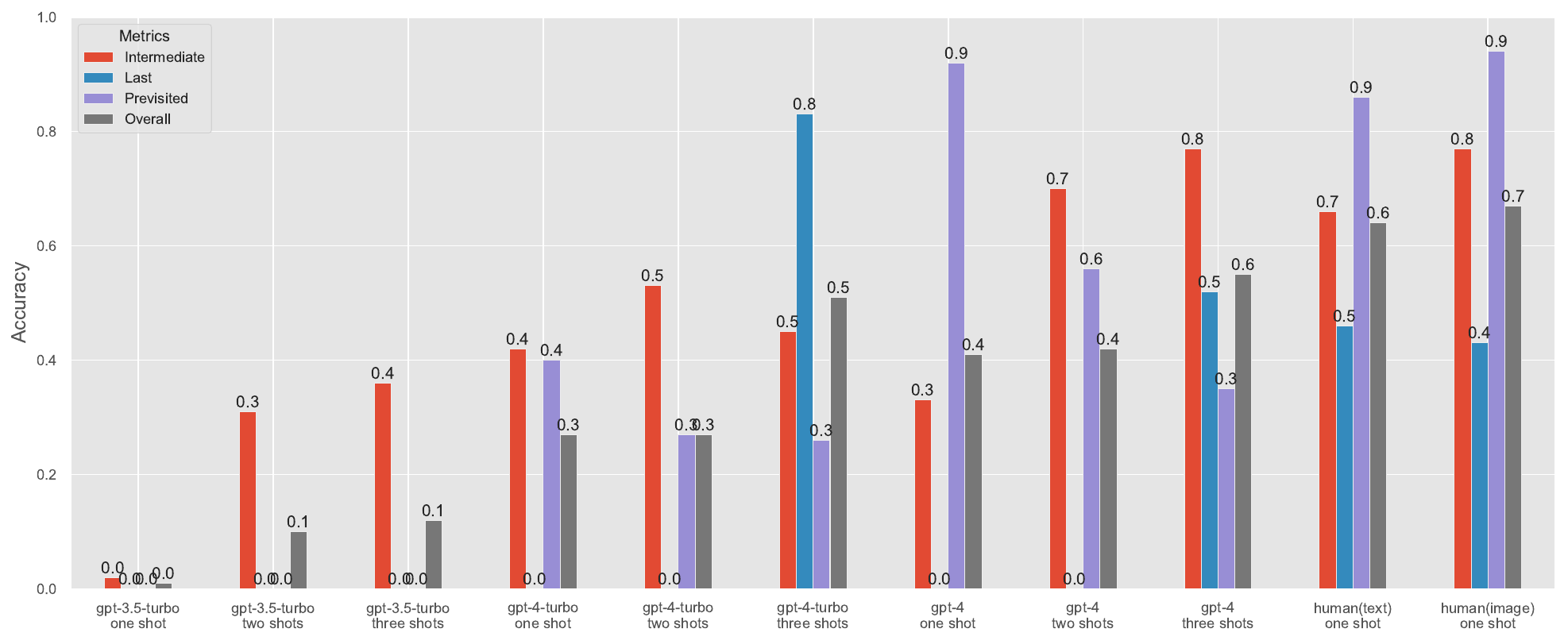}  
    }
\caption{\textbf{IR accuracy comparison on various few-shot cases.}}
\label{fig:FT various oneshot cases}
\end{figure*}

The in-context prompt contains at most 3 examples, denoted by 1-shot, 2-shot and 3-shot tests, respectively.
In the 1-shot test, only one fixed example of case \nTurn{} is inserted in the prompt before stating the testing \ac{ir} problem. In the 2-shot test, one fixed 
example of case \nDirect{} after one fixed example of case \nTurn{} are inserted in the prompt. In the 3-shot tests, three fixed examples, of case \nTurn, \nDirect, and \nFinal{}, are inserted in the prompt. As illustrated in \cref{fig:FT various oneshot cases}, all the models benefit from seeing examples, especially when examples in all cases are given.


\subsection{Shortcut Test}

Despite the cognitive nature of the tasks \ac{ir} and \ac{iip}, the task description and the answer appear in certain patterns.
It is possible that the language models we tested did not really use their ``cognitive capabilities'' (if they have) in answering those questions, but generating answers by recognizing the shortcut patterns instead.
It is difficult to confirm that the language models are making use of their cognitive capabilities, but much easier to see whether they learned to use certain shortcuts.
To explore this, we design experiments for both \ac{ir} and \ac{iip} to detect the presence of such shortcuts.

We conduct a test based on the previous \ac{ir} and \ac{iip} datasets. We neutralize the social and cognitive material as much as possible in description, which expose only the non-social part to \ac{llm}. The datasets are collections of neutralized \ac{ir} and \ac{iip} problems, each cut into a training set and a testing set, of volume ratio 5:1 (training vs testing). The training / testing sets are balanced to have the same distributions on types of problem. The model T5 is selected to learn the shortcuts via fine-tuning.




The shortcut version of \ac{ir} task is of generative form, given the modified prompt, the model T5 is requested to generate the preference pattern. In training (fine-tuning of T5), the data are the modified prompt-preference pattern pairs. 
For the modified \ac{ir} task prompt, we delete the description of question and setting, specifically ``campus'' and ``trajectories'', in order to avoid direct social and cognitive connections between output (preference) and the task context.

\begin{prompt}
*W*ZA*W****X****WWWM*Y***\\
(4, 0) view Z; memory Z \\ 
(4, 1) view Z; memory Z \\
(4, 2) view M; memory Z,M \\
... [Similarly all other intermediate points]\\
(3, 0) view Z; memory Z, M, X, Y; pick Z 
\end{prompt}


In the IIP task, we also removed detailed descriptions of the problem and background, extracting only the ``campus'' from each scenario and combining it with each of the four options to create individual samples for training and testing. This setup was designed as a generative task, where the model needed to identify the category of each option (\nMisld, \nShort, \nFar, \nBest) given a campus and an option. This methodology aimed to test the model's ability to understand and categorize options based on limited information.

\begin{prompt}
\texttt{W****W*W**WXW**W*W**Y**A*} \\
 + \textbf{Option 1}\\
\texttt{W****W*W**WXW**W*W**Y**A*} \\
 + \textbf{Option 2}\\
\texttt{W****W*W**WXW**W*W**Y**A*} \\
 + \textbf{Option 3}\\
\texttt{W****W*W**WXW**W*W**Y**A*} \\
 + \textbf{Option 4}
\end{prompt}

\subsection{GPT-4V Test}


Two question sets of volume 20 were selected from the \ac{ir} dataset and the \ac{iip} dataset, respectively.
The sets are used to conduct a batch-wise comparison of GPT-4V and humans on their abilities across multi-modal data for these two tasks, as shown in ~\Cref{tab:ir_gpt4v,tab:iip_gpt4v}. 
The statistics on the batch shows a potential that an extra image input results in a similar behavioral pattern for GPT-4V to that of GPT-4, compared based on the data in \cref{fig:ft_results} and \cref{fig:iip_stats}. By the time this task was performed, visual inputs to GPT-4V were only available on the OpenAI website, so we decided not to test GPT-4V on a larger dataset.


\begin{table}[H]
    \centering
    \caption{\textbf{Comparative Analysis of GPT-4V and Human Multimodal Abilities on IR.} We use accuracy (\%) as the metric.}
    \label{tab:ir_gpt4v}
        \begin{tabular}{lccc}
        \toprule
        & Favorite & Visible & Strict \\
        \midrule
        GPT-4V  & 0.65 & 0.60 & 0.20\\
        Human (image) & 0.75 & 0.70 & 0.60\\
        \bottomrule
        \end{tabular}%
\end{table}

\begin{table}[H]
    \centering
    \caption{\textbf{Comparative Analysis of GPT-4V and Human Multimodal Abilities on IIP.} Each row represents the distribution across four options.}
    \label{tab:iip_gpt4v}
        \begin{tabular}{lcccc}
        \toprule
        & \nShort{} & \nMisld{} & \nFar{} & \nBest{} \\
        \midrule
        GPT-4V  & 0.50 & 0.35 & 0.10 & 0.05\\
        Human (image) & 0.20 & 0.15 & 0.15 & 0.50\\
        \bottomrule
        \end{tabular}%
\end{table}

\end{document}